\documentclass{article}

\usepackage{PRIMEarxiv}

\usepackage[utf8]{inputenc} 
\usepackage[T1]{fontenc}    
\usepackage{hyperref}       
\usepackage{url}            
\usepackage{booktabs}       
\usepackage{amsfonts}       
\usepackage{nicefrac}       
\usepackage{microtype}      
\usepackage{lipsum}
\usepackage{fancyhdr}       
\usepackage{graphicx}       
\graphicspath{{media/}}     

\usepackage{amsmath}
\usepackage{physics}
\usepackage{makecell}
\usepackage{arydshln}
\usepackage{enumitem}
\usepackage{caption}
\usepackage{subcaption}
\usepackage{algorithm}
\usepackage{algpseudocode}
\newcommand{\overbar}[1]{\mkern 1.5mu\overline{\mkern-1.5mu#1\mkern-1.5mu}\mkern 1.5mu}
\usepackage{float}
\title{
Explainable Time Series Anomaly Detection using Masked Latent Generative Modeling}

\author{
  Daesoo Lee \\
  Norwegian University of Science and Technology \\
   \And
  Sara Malacarne \\
  Telenor Research \\
  \And
  Erlend Aune \\
  Norwegian University of Science and Technology \\ 
  BI Norwegian Business School \\
  HANCE \\ 
}

\begin{document}
\maketitle

\begin{abstract}
We present a novel time series anomaly detection method that achieves excellent detection accuracy while offering a superior level of explainability. Our proposed method, \textit{TimeVQVAE-AD}, leverages masked generative modeling adapted from the cutting-edge time series generation method known as TimeVQVAE. The prior model is trained on the discrete latent space of a time-frequency domain. Notably, the dimensional semantics of the time-frequency domain are preserved in the latent space, enabling us to compute anomaly scores across different frequency bands, which provides a better insight into the detected anomalies. Additionally, the generative nature of the prior model allows for sampling likely normal states for detected anomalies, enhancing the explainability of the detected anomalies through \textit{counterfactuals}. Our experimental evaluation on the UCR Time Series Anomaly archive demonstrates that TimeVQVAE-AD significantly surpasses the existing methods in terms of detection accuracy and explainability. 
We provide our implementation on GitHub: \url{https://github.com/ML4ITS/TimeVQVAE-AnomalyDetection}.
\end{abstract}


\section{Introduction}
Time series anomaly detection (TSAD) is a critical area of study in data analysis and machine learning, aiming to identify unusual patterns that deviate from expected behavior in fields such as finance, healthcare, and telecommunication. Various methods have been proposed for TSAD, leveraging techniques such as one-class classifications \cite{NIPS1999_8725fb77, ruff2018deep}, isolation forest \cite{liu2008isolation, guha2016robust}, discord discovery \cite{nakamura2020merlin, nakamura2023merlin++}, reconstruction \cite{audibert2021univariate, park2018multimodal, audibert2020usad, tuli2022tranad}, forecasting \cite{hundman2018detecting, harstad2021spatio}, and density estimation \cite{barz2018detecting, zong2018deep, dai2022graph}.
These methods have seemingly shown progress, primarily driven by deep learning methods. However, recent studies have exposed significant flaws in the popular benchmark datasets and evaluation protocols used in TSAD research \cite{wu2021current, rewicki2022worth, kim2022towards}. 
The benchmark datasets often suffer from unrealistic anomaly density and mislabeled ground truth \cite{wu2021current}. 
Additionally, these datasets may contain trivial problems that do not adequately assess the performance of TSAD methods. 
As for the evaluation protocol commonly used in TSAD, it is called Point Adjustment (PA), proposed by \cite{xu2018unsupervised}. However, this protocol has been criticized by \cite{kim2022towards} due to its potential to inflate performance results.
To address these issues, \cite{wu2021current} released the UCR Time Series Anomaly (UCR-TSA) archive that contains 250 curated benchmark datasets to provide more accurate evaluations. \cite{wu2021current} also suggested a simple scoring function to achieve a robust evaluation metric.
Since the introduction of the UCR-TSA archive in the literature, multiple studies \cite{rewicki2022worth, nakamura2023merlin++, mou2023deep} have evaluated existing TSAD methods using this benchmark. The findings of these studies were quite stunning, as they revealed that most of existing deep learning-based TSAD methods exhibit noticeably lower detection accuracies than their non-deep learning-based counterparts, contradicting their claims of state-of-the-art (SOTA) performance.
Furthermore, many papers plot few examples (as few as zero) even though time series analytics is inherently a visual domain \cite{wu2021current}.

\begin{figure}[!ht]
    \centering
    \includegraphics[width=0.65\textwidth]{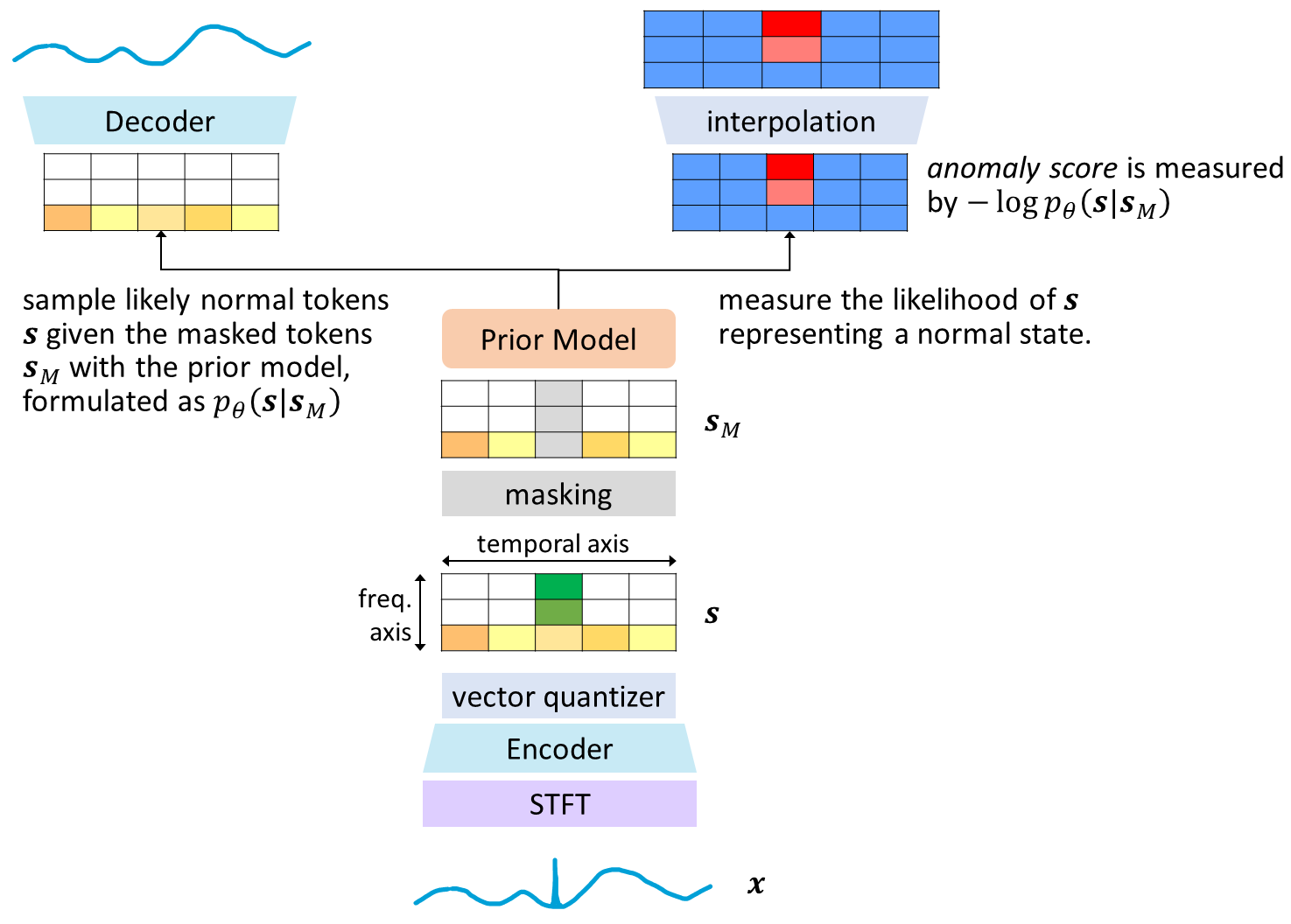}
    \caption{Overview of the inference process of our proposed method, TimeVQVAE-AD.
    In the figure, the time series $\textit{\textbf{x}}$ exhibits a high-frequency anomaly.
    Initially, $\textit{\textbf{x}}$ is processed by Short Time Fourier Transform (STFT), followed by processing through an encoder and a vector quantizer, resulting in $\textit{\textbf{s}}$. Here, $\textit{\textbf{s}}$ represents a set of tokens derived from $\textit{\textbf{x}}$, with color similarity indicating the Euclidean similarity between these tokens. The two axes of $\textit{\textbf{s}}$ correspond to time and frequency, respectively.
    Subsequently, a segment of $\textit{\textbf{s}}$ is masked to enable the prior model to sample likely normal tokens from the masked tokens (\textit{explainable sampling}), and to compute the anomaly scores (\textit{anomaly detection}) where low and high anomaly scores are depicted in blue and red, respectively.
    A noteworthy aspect of our method is its capability to address a broad spectrum of anomalies, due to the robust generative prior model that utilizes a learned prior to evaluate the likelihood of $\textit{\textbf{s}}$ representing a normal state.
    }
    \label{fig:overview_timevqvaead}
\end{figure}

In this paper, we present a novel TSAD method that achieves exceptional anomaly detection accuracy and offers a high level of explainability. 
The overview of the inference process of our method is illustrated in Fig.~\ref{fig:overview_timevqvaead}.
Unlike most of the existing deep methods for TSAD, our approach leverages masked generative modeling.
Masked generative modeling has demonstrated significant success in diverse fields, ranging from generative language modeling, such as BERT \cite{devlin2018bert}, to generative image modeling, such as DALL-E \cite{ramesh2021zero}.
In this work, we introduce a novel approach in which we predict anomaly scores directly from a learned prior using a robust prior model from TimeVQVAE \cite{lee2023vector}. TimeVQVAE is a SOTA time series generative method that has demonstrated superior performance in the literature \cite{ang2023tsgbench}.
Once the prior model is trained and learns the prior distribution using a training dataset, it can assign high probabilities to likely normal subsequences and low probabilities to abnormal subsequences of the time series. These probabilities can then be employed to calculate anomaly scores through the negative log-likelihood.
Our proposed method also benefits from TimeVQVAE for its explainability. The prior model of TimeVQVAE is trained on a time-frequency domain, rather than a time domain. This characteristic enables us to calculate anomaly scores across different frequency bands, consequently facilitating the factorization of anomalies in terms of anomaly types with respect to different frequency bands.
Additionally, presentation of likely normal states can be effortlessly accomplished by masking anomalous segments and conducting sampling using the learned prior model since the prior model is fundamentally generative. This allows us to approach explainable AI (XAI) for TSAD within the \textit{counterfactual} framework \cite{guidotti2022counterfactual, verma2020counterfactual}. The counterfactuals we generate verify, by definition, specific metrics adopted in this domain, such as 1) \textit{validity}, 2) \textit{plausibility} and 3) \textit{low computability time} \cite{guidotti2022counterfactual}.
Finally, we call our proposed method \textit{TimeVQVAE-AD}.

The experimental evaluation is performed on the UCR-TSA archive, and the experimental results demonstrate that TimeVQVAE-AD surpasses the performance of existing methods in terms of detection accuracy. Furthermore, it provides a superior level of explainability that, to the best of our knowledge, has not been achieved by any previous method. 
To boost confidence and transparency of our proposed method, we provide visualizations and CSV files of predicted anomalies for all 250 UCR-TSA archive datasets in our GitHub repository. 
This visualization evaluation protocol has been strongly recommended by \cite{wu2021current, nakamura2023merlin++} by pointing out at the issue that proper visualization practices have been significantly lacking in the literature despite the fact that TSAD belongs to the domain of time series analytics.

In summary, our contributions are
\begin{enumerate}[topsep=0pt,itemsep=-1ex,partopsep=1ex,parsep=1ex]
    \item[$\bullet$] TSAD via masked generative modeling,
    \item[$\bullet$] explainability via factorization of anomalies in terms frequency bands,
    \item[$\bullet$] explainability via counterfactuals,
    \item[$\bullet$] ground-breaking anomaly detection accuracy and explainability for TSAD,
    \item[$\bullet$] fair and robust evaluation on the UCR-TSA archive,
    \item[$\bullet$] availability of visualization for predicted anomalies across the UCR-TSA archive for transparency.
\end{enumerate}

\section{State of Time Series Anomaly Detection}
Over the years, numerous methods have been proposed for TSAD utilizing a range of mechanisms such as one-class classifications, isolation forest, discord discovery, reconstruction, forecasting, and density estimation.
Especially, a gradual improvement on the reported metrics has been observed, largely driven by the utilization of modern deep learning methods \cite{audibert2020usad, tuli2022tranad}. This progress has seemed to indicate that the field has been advancing. 

However, in recent years, a number of critical papers have emerged, highlighting significant flaws in the currently-popular benchmark datasets and evaluation protocol employed for TSAD methods \cite{wu2021current, rewicki2022worth, kim2022towards}. These papers reveal that the literature has been misled by inaccurately assessed metric scores, generating a false perception of progress while actual advancements have been minimal.

The recent study conducted by \cite{wu2021current} shed light on several critical issues regarding the currently-popular benchmark datasets employed to evaluate TSAD methods. The datasets that are commonly used now as benchmarks include Yahoo \cite{laptev2015s5}, Numenta \cite{ahmad2017unsupervised}, NASA \cite{hundman2018detecting}, and OMNI \cite{su2019robust}. \cite{wu2021current} argued that those datasets often suffer from mislabeled data, containing both false positives and false negatives, thereby leading to an inaccurate measure of detection accuracy. In addition, \cite{wu2021current} emphasized the presence of inconsistencies in labeling when dealing with similar instances of anomalies, resulting in an underestimation of the true positive rate for methods capable of detecting such anomalies. Furthermore, the study revealed that a substantial portion of the time series within those datasets can be easily solved through simple solutions, indicating that the trivial nature of the datasets fails to adequately assess the performance of TSAD methods. The flaws in these benchmark datasets have created a misleading perception of progress within the TSAD field, as methods that perform well on those datasets may not exhibit the same performance in real-world scenarios.
To address these issues, \cite{wu2021current} released a set of 250 carefully-curated benchmark datasets named UCR-TSA archive, where the datasets are free from these flaws and provide a more accurate evaluation of TSAD methods.

Another critical study \cite{kim2022towards} has pointed out the flaws of evaluating TSAD using the PA protocol. The idea behind PA is that if at least one moment in a contiguous anomaly segment is detected as an anomaly, the entire segment is then considered to be correctly predicted as an anomaly. The authors demonstrated that PA can lead to overestimation of the model's performance. For instance, even if the anomaly scores are randomly generated and cross the threshold only once within the ground truth segment, after applying PA, these predictions become indistinguishable from those of a well-trained model. This means that random anomaly scores can yield high F1 scores after PA, making it difficult to conclude that a model with a higher F1 score after PA performs better than others. 
To tackle the issue and establish a robust evaluation protocol, the author of the UCR-TSA archive suggested a simple yet fair evaluation protocol \cite{keogh2021multi}. 
During the evaluation process, it is expected that a method provides a singular predicted anomaly location. If this predicted location falls within a range of ±100 data points from the true location, it is considered correct, thereby achieving an accuracy of 1.0. On the other hand, if the predicted location deviates outside this range, it is considered incorrect, resulting in an accuracy of 0.0. Then the accuracies over the 250 datasets are averaged, and that is reported for the comparative evaluation.  

Experimental evidence supporting the flawed evaluation practices and the resultant misleading findings can be found in the literature \cite{nakamura2023merlin++, rewicki2022worth, mou2023deep}, where the experiments were conducted on popular deep TSAD methods. For instance, TranAD \cite{tuli2022tranad} is a recent deep TSAD method built on a transformer model \cite{vaswani2017attention} and it claimed to be SOTA for TSAD. In its paper, TranAD outperformed its 10 competing methods, achieving F1 score of 0.94 on the Numenta dataset and 0.89 on the NASA dataset. 
However, when evaluated on the UCR-TSA archive, TranAD exhibited remarkably poor performance, achieving the averaged accuracy of merely 0.16 -- even worse than the autoencoder (AE)-based method that achieved an accuracy of 0.28 \cite{audibert2021univariate}.

\begin{figure}
    \centering
    \begin{subfigure}[b]{0.47\textwidth}
        \centering
        \includegraphics[width=\textwidth]{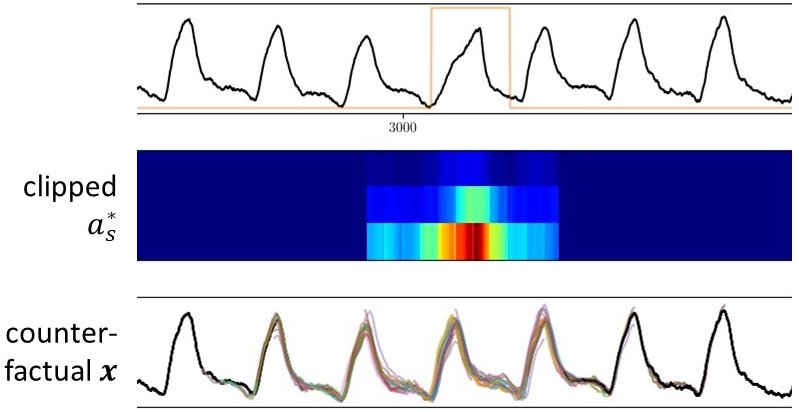}
        \caption{LF anomaly}
    \end{subfigure}
    \hspace{0.02\textwidth}
    \begin{subfigure}[b]{0.47\textwidth}
        \centering
        \includegraphics[width=\textwidth]{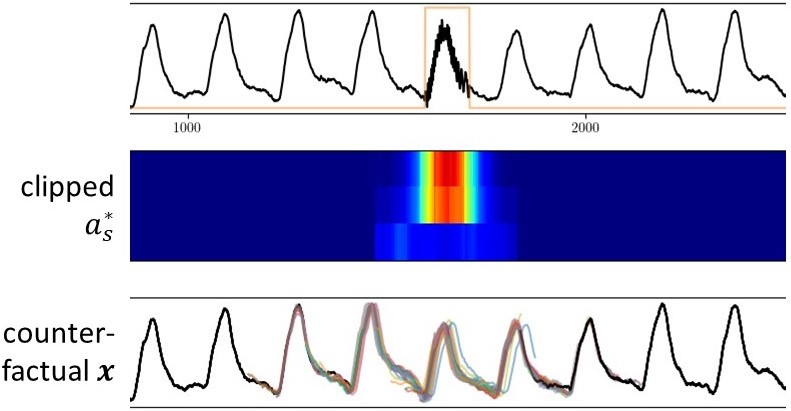}
        \caption{HF anomaly}
    \end{subfigure}
    \caption{Example of the two perspectives of explainability: 1) factorization of anomalies in terms of anomaly types via frequency decomposition, 2) presentation of a corresponding normal state. 
    The two subfigures show time series with different anomaly types: LF and HF anomalies, respectively.
    In each subfigure, the first figure presents a time series (black) with anomaly labels (orange), the second figure presents predicted anomaly scores with respect to different frequency bands using our proposed method, where the bottom row and the top row represent the lowest and highest frequency band, respectively (blue: low anomaly score, red: high anomaly score), and the third figure shows the likely normal states of the time series, in which the anomalous segments are resampled using our learned prior model.
    Note that the predicted anomaly scores are high in the low frequency band in (a), the scores are high in the high frequency band in (b), and the likely normal states are highly convincing when observed by human eyes.}
    \label{fig:examples_LF_HF_anomaly_detection}
\end{figure}

One additional limitation we have observed in the existing methods is the absence of explainability. Given that TSAD operates within the domain of time series analytics, it is essential for models to provide users with diagnostic explanations. These explanations play a critical role in delivering valuable insights about the detected anomalies, thereby enabling users to gain a deeper understanding of the underlying reasons behind them.
In our work, we provide two important perspectives for XAI for TSAD: 1)  \textit{factorization of anomalies in terms of anomaly types}, 2) \textit{presentation of likely normal realizations, i.e., counterfactuals}.
As for the first perspective, there exists a diverse range of anomalies in time series data, including local peaks, noise, steep increases, signal shifts, unusual patterns, and more \cite{rewicki2022worth}. These different types of anomalies can often be categorized based on their frequency characteristics. For instance, local peaks and i.i.d noise can be considered high-frequency (HF) anomalies, while signal shifts and drifts can be classified as low-frequency (LF) anomalies. Additionally, unusual patterns can exhibit anomalous behavior in both LF and HF frequency bands. 
This anomaly type factorization in terms of frequency offers better interpretability of the detected anomalies, 
as depicted in Fig.~\ref{fig:examples_LF_HF_anomaly_detection}, in which a HF anomaly may indicate a sensor noise and a LF anomaly can indicates a systematic change in the measured subject.
The second perspective of explainability involves providing insights into what the data would look like in its non-anomalous state -- counterfactuals in the XAI literature \cite{guidotti2022counterfactual}. For example, a patient diagnosed with a brain tumor can better comprehend the doctor's assessment by comparing their brain image to that of a normal brain \cite{marimont2021anomaly}. Similarly, this principle can be applied to TSAD, as depicted in Fig.~\ref{fig:examples_LF_HF_anomaly_detection}. By offering a comparison between an anomalous time series and corresponding likely realizations over the anomaly window, an enhanced explainability of the detected anomalies can be achieved. 

\section{Related Work}

\subsection{Existing Anomaly Detection Methods}
The existing methods for TSAD can be categorized into two groups: 1) Non-Deep Learning (DL)-based methods, and 2) DL-based methods.

\subsubsection{Non-Deep Learning-based TSAD Methods} 
Non-DL-based TSAD methods utilize various mechanisms, including one-class classification \cite{scholkopf1999support, ruff2018deep}, isolation forest \cite{liu2008isolation, guha2016robust}, density estimation \cite{barz2018detecting}, matrix profile \cite{yeh2016matrix, law2019stumpy}, and discord discovery \cite{nakamura2020merlin, nakamura2023merlin++}. Among these approaches, matrix profile techniques such as SCRIMP \cite{yeh2016matrix} and discord discovery methods such as MERLIN++ \cite{nakamura2023merlin++} demonstrate the highest detection accuracy. Furthermore, MERLIN++ offers greater computational efficiency.
Both matrix profile and discord discovery methods calculate anomaly scores by measuring the distances between different subsequences of time series data. However, there is a difference in their computational requirements. Matrix profile methods can be computationally intensive due to the need to calculate pairwise distances between all pairs of subsequences. On the other hand, discord discovery focuses on identifying subsequences with large pairwise distances, resulting in a significant reduction in computational cost.

\subsubsection{Deep Learning-based TSAD Methods}
\begin{figure}[h]
    \centering
    \begin{subfigure}[b]{0.47\textwidth}
        \centering
        \includegraphics[width=\textwidth]{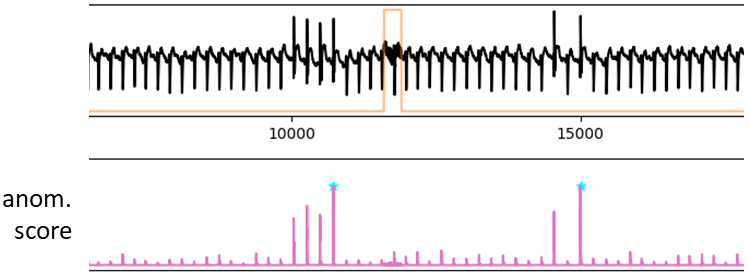}
        \caption{anomaly with small amplitude (noise)}
    \end{subfigure}
    \hspace{0.02\textwidth}
    \begin{subfigure}[b]{0.47\textwidth}
        \centering
        \includegraphics[width=\textwidth]{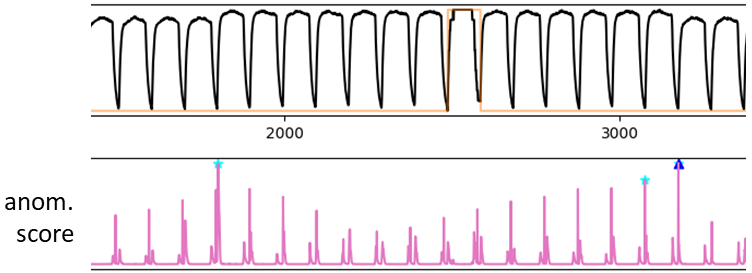}
        \caption{anomaly with subtle pattern difference}
    \end{subfigure}
    \caption{Examples of inevitable failure cases for reconstruction or forecasting-based TSAD methods. In each subfigure, the second figure presents predicted anomaly scores (pink). 
    A reconstruction or forecasting error $\| x_\mathrm{train} - \hat{x} \|$ tends to be larger on timesteps with large amplitudes, as they are more challenging to predict. This results in higher anomaly scores at peaks. In contrast, anomalies with small amplitudes inevitably yield low predicted anomaly scores due to their nature, as observed in (a) and (b).
    }
    \label{fig:examples_autoencoder_failure_cases}
\end{figure}

With the advancements in DL, many approaches for TSAD based on DL have emerged. The most common existing DL methods are based on reconstruction or forecasting tasks \cite{malhotra2016lstm, hundman2018detecting}. Additionally, there have been attempts based on different approaches such as adversarial training \cite{li2019mad, geiger2020tadgan, saravanan2023tsi}, density estimation \cite{zong2018deep, dai2022graph}, and (non-)contrastive learning \cite{de2021contrastive, mou2023deep}.
Moreover, some recent studies such as \mbox{\cite{shin2023time,fu2022mad}} adopted masked modeling with anomaly score measurement based on reconstruction. Although these studies utilize masked modeling, their approaches and our approach are fundamentally different in terms of anomaly detection because their anomaly scores are measured based on reconstruction while ours on density estimation.
However, the critical papers by \cite{rewicki2022worth, nakamura2023merlin++} have demonstrated that the DL-based TSAD methods do not meaningfully outperform non-DL methods such as SCRIMP, MERLIN \cite{nakamura2020merlin}, and MERLIN++ when fairly evaluated on the UCR-TSA archive.
We have identified the intrinsic limitations of DL-based methods as follows:

\paragraph{Limitation of Reconstruction or Forecasting-based TSAD Methods} 
The reconstruction or forecasting-based methods are trained by minimizing the error $\| x_\mathrm{train} - \hat{x} \|$, where $x_\mathrm{train}$ represents the model input during training and $\hat{x}$ represents the model output for reconstruction or forecasting. Once the model is properly trained and well regularized, it should perform well on reconstructing $x_\mathrm{test}$, resulting in a small test error and equivalently, a small anomaly score. Anomalies with unusually-high amplitude can be easily captured since their errors will be large due to the large amplitude of $x_\mathrm{test}$. However, these methods inherently struggle to detect anomalies with small amplitude or subtle pattern differences, as depicted in Fig.~\ref{fig:examples_autoencoder_failure_cases}. Experimental evidence aligned with this claim is provided in \cite{rewicki2022worth}.

\paragraph{Limitation of Adversarial Learning-based TSAD Methods}
The adversarial learning or Generative Adversarial Network-based methods involve training a generator and discriminator, where the discriminator is trained to distinguish between real and generated samples. These methods measure anomaly scores by utilizing the discriminator score. However, using the discriminator score as an anomaly score can lead to potential issues, as the discriminator is primarily designed to distinguish between real and generated samples rather than specifically identifying anomalies -- that is, there exists a definitive distinction between fake samples and anomalous samples.  
Furthermore, due to mode collapse in the generator, it is unlikely to generate a diverse range of anomalous patterns. As a result, the discriminator may not be able to accurately assign discriminator scores to unseen anomalous patterns.

\paragraph{Limitation of Existing Density Estimation-based TSAD Methods}
The density estimation-based methods aim to measure anomaly scores via a learned prior distribution. DAGMM \cite{zong2018deep}, which stands for Deep Autoencoding Gaussian Mixture Model, is a highly-cited method for anomaly detection, and \cite{bhatnagar2021merlion} adopted it for time series. DAGMM consists of a compression network and an estimation network. The compression network compresses a sample into a latent vector, and the estimation network performs density estimation under the framework of Gaussian Mixture Modeling (GMM). The anomaly score is computed based on the likelihood of the latent vector in the Gaussian distribution modeled by GMM. However, DAGMM faces a similar challenge as the reconstruction or forecasting-based methods, as the latent vector obtained with the compression network retains the shape information of a sample, making it likely that similarly-shaped time series are located close in the Gaussian distribution \cite{spinner2018towards}. Consequently, it is prone to failing in capturing anomalies with small amplitude or subtle pattern difference. Another recent proposal, GANF (Graph Augmented Normalizing Flow), measures anomaly scores based on its learned prior using normalizing flow. 
However, this approach is inherently limited by the properties of normalizing flow. 
\cite{kirichenko2020normalizing} showed the failure of normalizing flows in detecting out-of-distribution data and investigated the reasons behind it. They found that the failure can be attributed to the model's inability to learn high-level semantic features and its shortcomings in global structure learning.

While most TSAD methods operate within a time domain, a study conducted by \mbox{\cite{zhang2022tfad}} proposed measuring anomaly scores across both time and frequency domains while using a distance metric for anomaly score measurement. The utilization of a frequency domain allows for capturing anomalous time series with a small magnitude in a time domain and a high magnitude in a frequency domain, at which the reconstruction-based methods fail. However, it only partially resolves the limitation of the reconstruction-based methods because anomalies can only be captured when they carry a high magnitude either in a time or frequency domain due to the nature of a distance metric. Consequently, this limitation hinders its ability to detect subtle anomalies.

\subsection{TimeVQVAE: A Powerful Time Series Generation Method}
TimeVQVAE introduces a novel approach for time series generation, inspired by the success of masked image generative modeling \cite{van2017neural, yu2021vector, yu2022scaling}. 
TimeVQVAE focuses on producing high-quality synthetic samples through two key stages: tokenization (stage~1) and prior learning (stage~2). In the tokenization stage, TimeVQVAE utilizes vector quantization modeling in the time-frequency domain. This allows for effective representation and encoding of time series data. The subsequent prior learning stage involves training transformer models to learn a prior distribution of time series data. Notably, TimeVQVAE exhibits superior performance compared to other methods in both unconditional and class-conditional sampling, signifying a significant advancement in the field of time series generation.
Because TimeVQVAE can learn a prior distribution, it can also perform anomaly detection, similar to DAGMM and GANF. However, unlike DAGMM and GANF, TimeVQVAE does not impose any bias on the shape of the prior distribution and learns the prior via masked generative modeling with a bidirectional transformer model. By doing so, TimeVQVAE overcomes the challenge of prior learning and can produce a learned prior distribution that robustly estimates a target prior distribution. 
In this work, we harness the powerful prior learning of TimeVQVAE and tailor its usage for anomaly detection, while incorporating explainability through its inherent generative nature.

\subsection{Existing Approaches for XAI for TSAD}

In \cite{sulem2022diverse}, they adopt diverse counterfactuals for explaining anomalies in time series, and argue that a diverse set of counterfactuals are desirable to get the right intuition for why a detected anomaly is indeed anomalous. They elaborate on the importance of visualization for TSAD. In \cite{tripathy2022explaining}, they survey XAI for TSAD, focusing on multivariate industrial time series. They mention LIME and SHAP - two classical methods for feature attribution that may help explain how covariates contribute to an anomaly. In their review, they summarize the pros and cons of methods in a table (Table 1 in \cite{tripathy2022explaining}). Common for the methodologies is the use of visualizations to explain an anomaly. For a general overview the taxonomy, opportunities and limitation of XAI, we refer to the survey \cite{arrieta2020explainable}.

\section{Method}
Before diving into the details of our method, it is important to establish the space in which the prior distribution resides and introduce relevant notations.
TimeVQVAE follows the two-stage training approach by VQ-VAE \cite{van2017neural}. In the first stage, an encoder $E$, vector quantizer $VQ$, and decoder $D$ are trained by minimizing a reconstruction loss. In the context of TimeVQVAE, a given time series $\textit{\textbf{x}}$ undergoes a STFT preprocessing step and is subsequently encoded and quantized into a discrete latent vector $\textit{\textbf{z}}_q$, expressed as $\textit{\textbf{z}}_q = VQ(E(\mathrm{STFT}(x)))$ where $\textit{\textbf{x}} \in \mathbb{R}^{T}$, $\mathrm{STFT}(\textit{\textbf{x}}) \in \mathbb{R}^{2 \times H \times T^\prime}$, and $\textit{\textbf{z}}_q \in \mathbb{R}^{D \times H \times W}$.
$T$ denotes the time series length, $2$, $H$, and $T^\prime$ denotes real and imaginary channels, the frequency dimension (height), and the temporal length of $\mathrm{STFT}(\textit{\textbf{x}})$, respectively, and $D$ and $W$ denote the latent dimension size and the latent temporal length (width), respectively.
Next, the second stage utilizes MaskGIT \cite{chang2022maskgit} to learn the prior of the discrete latent space $p(\textit{\textbf{z}}_q)$ via masked modeling with a bidirectional transformer model (\textit{i.e.,} prior model), and the learned prior distribution is denoted as $p_\theta(\textit{\textbf{z}}_q)$ where $\theta$ denotes parameters of the prior model. 
In the following, we use the notation of tokens $\textit{\textbf{s}} \in \mathbb{R}^{H \times W}$ instead of $\textit{\textbf{z}}_q \in \mathbb{R}^{D \times H \times W}$ for brevity, where a token $s$ refers to the codebook index of $z_q$, where $s \in \textit{\textbf{s}}$ and $z_q  \in \textit{\textbf{z}}_q$ \cite{chang2022maskgit}. Using the notation, the target prior distribution is expressed as $p(\textit{\textbf{s}})$ instead of $p(\textit{\textbf{z}}_q)$.


The prior distribution $p(\textit{\textbf{s}})$ is learned by $p_\theta(\textit{\textbf{s}})$ via masked modeling. Masked modeling is often adopted for language modeling \cite{devlin2018bert}, self-supervised learning \cite{baevski2022data2vec}, and generative modeling \cite{chang2022maskgit}, and its objective is to maximize $p_\theta(\textit{\textbf{s}} | \textit{\textbf{s}}_M)$ where $\textit{\textbf{s}}_M$ denotes a masked version of $\textit{\textbf{s}}$ and is defined as $\textit{\textbf{s}}_M = \textit{\textbf{s}} \odot \textit{\textbf{m}} + \texttt{[MASK]} \odot (1 - \textit{\textbf{m}})$ where $\textit{\textbf{m}}$ consists of 0 and 1 and \texttt{[MASK]} represents a mask token. 
In the context of natural language processing, $\textit{\textbf{s}}$ and $\textit{\textbf{s}}_M$ can be $[\texttt{apple}, \texttt{is}, \texttt{red}]$ and $[\texttt{apple}, \texttt{is}, \texttt{[MASK]}]$, respectively, and the prior model can calculate the probability of $\texttt{red}$ given the other words $\texttt{apple}$ and $\texttt{is}$. 
The same approach can be applied to time series. For instance, TimeVQVAE adopted this masked modeling approach for time series generation and achieved SOTA performance. 
We redirect the use of the learned prior model of TimeVQVAE to perform TSAD instead of generation. In the training of masked modeling, 
a uniform-random portion of $\textit{\textbf{s}}$ is uniform-randomly masked, therefore the masking can be flexible in terms of size and locations during inference. 
Then, we can have a sliding masking window with an arbitrary window size and mask $\textit{\textbf{s}}$ and measure $p_\theta(\textit{\textbf{s}} | \textit{\textbf{s}}_M)$ iteratively along the temporal dimension. 
Finally, our anomaly score for the masked region $M^\prime$ can be calculated as $\textit{\textbf{a}}_{M^\prime} = \textit{\textbf{a}} \odot (1 - \textit{\textbf{m}})$ where $\textit{\textbf{a}} = - \log{p_\theta(\textit{\textbf{s}} | \textit{\textbf{s}}_M)}$ and $\textit{\textbf{a}}_{M^\prime}$ is the predicted anomaly score for the masked region in $\textit{\textbf{s}}_M$. 

As $p_\theta(\textit{\textbf{s}} | \textit{\textbf{s}}_M)$ is a crucial component in our TSAD, we delve deeper into the inner workings of it: a process of assigning higher probabilities to likely states of time series and lower probabilities to less and unlikely states.

The anomaly scores are measured using $p_\theta(\textit{\textbf{s}} | \textit{\textbf{s}}_M)$ and each element of the probability can be represented in the softmax form as 
\begin{subequations}
\begin{align} 
p_\theta(\textit{\textbf{s}}_i | \textit{\textbf{s}}_M) &= \frac{e^{(\textit{\textbf{u}}_i)_{k^*}}}{\sum_{k=1}^K {e^{ (\textit{\textbf{u}}_i)_k }}}, \\
\textit{\textbf{u}} &= f_\theta(\textit{\textbf{s}}_M),
\end{align}
\end{subequations}
where the subscript $i$ denotes an arbitrary element within the spatial dimension, $k$ and $K$ denote a codebook index and codebook size, respectively, $f_\theta$ denotes the prior model, $\textit{\textbf{u}} \in \mathbb{R}^{H \times W \times K}$, $\textit{\textbf{u}}_i \in \mathbb{R}^{K}$, and $(\textit{\textbf{u}}_i)_{k^*}$ denotes a prediction of $u$ for $\textit{\textbf{s}}_i$, where $k^*$ denotes a codebook index of $\textit{\textbf{s}}_i$.
In the second stage, the prior model is trained by maximizing $p_\theta(\textit{\textbf{s}}_i | \textit{\textbf{s}}_M)$, leading to assigning higher values for $\textit{\textbf{u}}_i$ indexed by $k^*$ while assigning lower values for $\textit{\textbf{u}}_i$ not indexed by $k^*$. The former corresponds to producing higher probabilities for likely normal states and the latter corresponds to producing lower probabilities for less likely and unlikely states.
As a result, during inference, when given a time series containing an anomaly, the prior model can produce a low probability for the anomaly by performing the masked prediction with the mask on the anomaly. 

In addition, $p_\theta(\textit{\textbf{s}} | \textit{\textbf{s}}_M)$ indicates that it is a stochastic generative process. 
As illustrated in Fig.~\ref{fig:overview_timevqvaead}, when given a token set with an anomalous segment masked, the prior model can stochastically generate likely normal states of the tokens.
This enables the sampling of likely normal states of time series (counterfactual examples).


\subsection{Training}
\begin{figure}[!ht]
    \centering
    \begin{subfigure}[b]{0.49\textwidth}
        \centering
        \includegraphics[width=\textwidth]{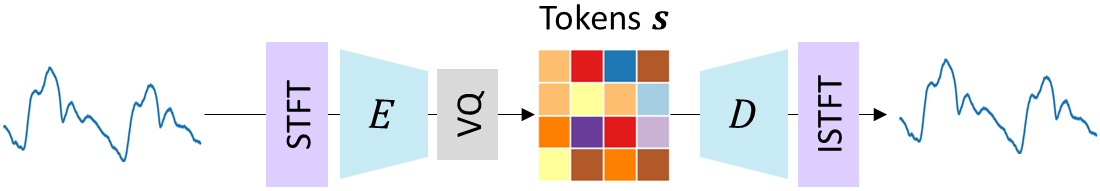}
        \caption{stage~1}
    \end{subfigure}
    \hfill
    \begin{subfigure}[b]{0.49\textwidth}
        \centering
        \includegraphics[width=\textwidth]{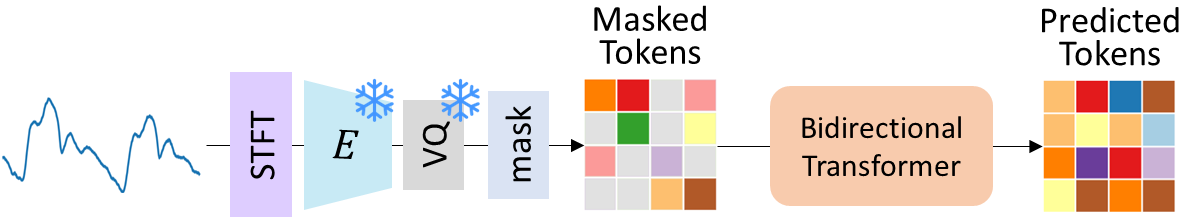}
        \caption{stage~2}
    \end{subfigure}
    \caption{Overview of the first stage (stage~1) and the second stage (stage~2).
    The snowflakes indicate that the models are set to be untrainable, \texttt{mask} denotes random-masking, and the bidirectional transformer corresponds to the prior model. 
    }
    \label{fig:overview_stage12}
\end{figure}

Our prior model is an architectural variant of TimeVQVAE, therefore the training process remains the same. TimeVQVAE adopts the two-stage training approach from VQ-VAE.  
Mathematically, the first stage is formulated as minimizing 
\begin{align} 
\label{eq:}
\mathbb{E}_{\textit{\textbf{x}} \sim \textit{\textbf{X}}} \| \textit{\textbf{x}} - D(VQ(E(\mathrm{STFT}(\textit{\textbf{x}})))) \|
\end{align}
and the second stage is formulated as maximizing
\begin{align}
\label{eq:}
\mathbb{E}_{\textit{\textbf{x}} \sim \textit{\textbf{X}}, \textit{\textbf{z}}_q \sim VQ(E(\mathrm{STFT}(\textit{\textbf{x}})))} \left[ p_\theta(\textit{\textbf{s}} | \textit{\textbf{s}}_M) \right],
\end{align}
where $E$ and $D$ are trained in the first stage and set to be untrainable in the second stage. 
Fig.~\ref{fig:overview_stage12} illustrates the overview of the first and second stages.

\subsection{Architecture}
\label{sect:architecture}
We propose the two architectural modifications to TimeVQVAE: 1) \textit{LF-HF latent space merge}, 2) \textit{dimensional semantics-preservative convolutional encoder}. 

\paragraph{LF-HF Latent Space Merge}
In TimeVQVAE, two sets of $E$, $VQ$, and $D$ are used for LF and HF, respectively, to allow high compression on the LF latent space, leading to enhanced sampling performance. 
With the small LF latent space, the prior modeling $p_\theta(\textit{\textbf{s}} | \textit{\textbf{s}}_M)$ becomes easier, so as the sampling process $p_\theta(\textit{\textbf{s}}_{M,0:T_s}) = \prod_{t_s=0}^{T_s-1} p(\textit{\textbf{s}}_{M,t_s+1} | \textit{\textbf{s}}) p_\theta(\textit{\textbf{s}} | \textit{\textbf{s}}_{M,t_s})$, in which the recursive product represents the iterative decoding proposed by \cite{chang2022maskgit}, $\textit{\textbf{s}}_{M,0}$ is equivalent to a complete set of mask tokens $\texttt{[MASK]}$, and $\textit{\textbf{s}}_{M,T_s} = \textit{\textbf{s}}_{T_s}$ is equivalent to a set of fully-sampled tokens. 
We, however, do not need to solve $p_\theta(\textit{\textbf{s}}_{M,0:T_s})$ for anomaly detection. 
In cases where a time series exhibits anomalies, it is common to find that only a specific segment of the series is affected. Consequently, a particular segment of $\textit{\textbf{s}}$ carry tokens representing anomalous states.
Therefore, we only need to solve a partial problem of $p_\theta(\textit{\textbf{s}}_{M,0:T_s})$ as $p_\theta(\textit{\textbf{s}}_{M,t:T})$ where $t > 0$  to compute the anomaly score of a certain-sized masked region $\textit{\textbf{a}}_{M^\prime}$. 
Consequently, $p_\theta(\textit{\textbf{s}}_{M,t:T})$ is an easier problem on its own and we no longer need to pose the high compression on the LF latent space.
This allows us to merge the LF and HF latent spaces into a unified latent space. This not only simplifies the architecture but also enables us to capture fine-grained anomalies since $s$ or $z_q$ has a smaller receptive field due to the smaller compression rate. As a result, we only need a single set of $E$, $VQ$, and $D$ instead of two sets, as shown in Fig.~\ref{fig:overview_stage12}. 

\paragraph{Dimensional Semantics-preservative Convolutional Encoder}
Our anomaly scores are computed in the discrete latent space using our learned prior using $\textit{\textbf{a}} = -\log{p_\theta(\textit{\textbf{s}} | \textit{\textbf{s}}_M)}$. Hence, it is crucial to preserve the semantics of both the temporal and frequency dimensions of $\mathrm{STFT}(\textit{\textbf{x}}) \in \mathbb{R}^{2 \times H \times T^\prime}$ so that the predicted anomaly scores are temporally aligned with $\textit{\textbf{x}}$ while preserving the distinct frequency bands.
However, the current configuration of the residual blocks in the encoder of TimeVQVAE, which use a kernel size of $\texttt{(3x3)}$, and the downsampling blocks, which use a kernel size of $\texttt{(3x4)}$, causes wide-mixture of information across the temporal and frequency dimensions. This wide-mixture significantly loosens the semantics of both dimensions.
To address this limitation, we introduce a dimensional semantics-preservative convolutional encoder. This encoder is similar to the one used in TimeVQVAE, but with a key modification: the kernel sizes. 
Specifically, we propose replacing the kernel sizes of the residual block and downsampling block with $\texttt{(1x3)}$ and $\texttt{(1x4)}$, respectively. By making this adjustment, we ensure that $\textit{\textbf{z}}_q$ and its corresponding tokens $\textit{\textbf{s}}$ remain independent along the frequency axis and preserve sharper temporal semantics.
Importantly, the decoder should also have the frequency-independent kernel sizes (\textit{i.e.,} $\texttt{(1x4)}$). Otherwise, the decoder learns to predict $i$-th frequency information utilizing $j$-th frequency information, where $i \neq j$, allowing the encoder not to strictly encode the $i$-th frequency information, leading to ineffective discrete latent space modeling with respect to different frequencies, thereby sub-optimal performance.

\subsection{Anomaly Score Prediction Process}
Our anomaly scores reside in the discrete latent space. Yet, because the proposed convolutional encoder preserves the semantics of temporal and frequency dimensions, we can easily map the anomaly scores in the discrete latent space to the time series data space. 
We first discuss the anomaly score prediction in the discrete latent space and then mapping the scores onto the data space.
Pseudocode for the anomaly score prediction process is presented in Algorithm~\ref{alg:anomaly_score_prediction}.

\subsubsection{Anomaly Score Prediction in the Discrete Latent Space}
To systematically measure the anomaly scores, we propose to compute $\textit{\textbf{a}} = -\log{p_\theta(\textit{\textbf{s}} | \textit{\textbf{s}}_M)}$ using a \textit{sliding masking latent window} along the temporal dimension. The masking latent window refers to masking a certain-sized temporal segment across the frequency dimension in the discrete latent space, and the sliding refers to the repetition of the latent window-masking process along the temporal dimension.  
To be precise, for $\textit{\textbf{s}} \in \mathbb{R}^{H \times W}$ and the certain temporal step $w$, we compute a
predicted anomaly score as
\begin{align}
\textit{\textbf{a}}_w = \mathbb{E}_w [ -\log{p_\theta(\textit{\textbf{s}}_{:, w-\alpha : w+\alpha} | s_{M(:, w-\alpha : w+\alpha)})} ],
\end{align}
where $\textit{\textbf{a}}_w \in \mathbb{R}^H$, $-\log{p_\theta(\textit{\textbf{s}}_{:, w-\alpha : w+\alpha} | s_{M(:, w-\alpha : w+\alpha)})} \in \mathbb{R}^{H \times 2\alpha}$, the subscript $[ :, w-\alpha:w+\alpha ]$ indicates indexing the specified temporal (width) range from $w-\alpha$ to $w+\alpha$ across the frequency dimension $H$, $\alpha$ is an positive integer, and $M(:, w-\alpha:w+\alpha)$ indicates masking the specified temporal segment across the frequency dimension.

We adopt $-\log{p_\theta(\textit{\textbf{s}}_{:, w-\alpha : w+\alpha} | s_{M(:, w-\alpha : w+\alpha)})}$ instead of $-\log{p_\theta(\textit{\textbf{s}}_{:, w} | s_{M(:, w-\alpha : w+\alpha)})}$ because the latent temporal step $w$ shares the information with the neighboring steps due to the receptive field of the convolutional encoder and we experimentally found that the former results in more robust anomaly score prediction.

We repeat the process along the temporal dimension $W$ using a sliding latent window for all $w$, and obtain $\tilde{\textit{\textbf{a}}} \in \mathbb{R}^{H \times W}$ which is aligned with $\mathrm{STFT}(\textit{\textbf{x}}) \in \mathbb{R}^{H \times T^\prime}$ in terms of the dimensional semantics.
We can obtain multiple $\tilde{\textit{\textbf{a}}}$ with a set of different $\alpha$-s to incorporate the predicted anomaly scores with different latent window sizes, as $\alpha$ analogically specifies a kernel size. 
Then the multiple $\tilde{\textit{\textbf{a}}}$ can be summed to combine the effects of different latent window sizes.

\begin{algorithm}[!t]
\caption{Pseudocode of the anomaly score prediction process using the learned prior model}
\label{alg:anomaly_score_prediction}
\begin{algorithmic}
\State \textbf{Define} a set of different $\alpha$-s $\{\alpha_0, \alpha_1, ...\}$
\State \textbf{Define} an entire sequence of time series $\textit{\textbf{x}}^* \in \mathbb{R}^{T^*}$
\State \textbf{Define} a period length $P$ of $\textit{\textbf{x}}^*$
\State \textbf{Define} multiple $\textit{\textbf{a}}^* \in \mathbb{R}^{H \times T^*}$ for $\{\alpha_0, \alpha_1, ... \}$, initialized with zeros.

\\
\For{$\alpha \in \{\alpha_0, \alpha_1, ...\}$}  \Comment{This can be replaced with parallel computation via multi-processing}
\For{$t \in [0, 1, 2, ...]$}  \Comment{stride can be applied for faster computation}
    \State $\textit{\textbf{x}} \gets \textit{\textbf{x}}^*_{t:t + T}$  \Comment{$\textit{\textbf{x}} \in \mathbb{R}^{T}$, $T = 2P$, $[:, t: t + T]$ indexes $\textit{\textbf{x}}^*$ from $t$ to $T$}
    \State $\textit{\textbf{z}}_q \gets VQ(E(\mathrm{STFT}(\textit{\textbf{x}})))$  \Comment{$\textit{\textbf{z}}_q \in \mathbb{R}^{D \times H \times W}$}
    \State $\textit{\textbf{s}} \gets$ the codebook indices of $\textit{\textbf{z}}_q$  \Comment{$\textit{\textbf{s}} \in \mathbb{R}^{H \times W}$}

        \For{$w \in [0, 1, ..., W]$} 
            \State $\textit{\textbf{a}}_w \gets \mathbb{E}_{w} \left[ -\log{p_\theta( \textit{\textbf{s}}_{:,w-\alpha:w+\alpha} | \textit{\textbf{s}}_{M(:,w-\alpha:w+\alpha)} )} \right]$  \Comment{$\textit{\textbf{a}}_w \in \mathbb{R}^{H}$ denotes the predicted anomaly score at $w$}
            \State Store $\textit{\textbf{a}}_w$
        \EndFor
        \State $\tilde{\textit{\textbf{a}}} \gets$ a collection of $\textit{\textbf{a}}_w$  \Comment{$\tilde{\textit{\textbf{a}}} \in \mathbb{R}^{H \times W}$}

    \State $\tilde{\textit{\textbf{a}}}_m \gets$ mapping $\tilde{\textit{\textbf{a}}}$ to the data space with simple nearest-neighbor interpolation  \Comment{$\tilde{\textit{\textbf{a}}}_m \in \mathbb{R}^{H \times T}$}
    \State{${\textit{\textbf{a}}}^*_{:, t: t + T} = {\textit{\textbf{a}}}^*_{:, t: t + T} + \tilde{\textit{\textbf{a}}}_m$}  \Comment{$[:, t: t + T]$ indexes $\textit{\textbf{a}}^*$ from $t$ to $T$ across the frequency dimension}
    
\EndFor

\EndFor

\State $\textit{\textbf{a}}^{*}_s \gets$ a summation of multiple $\textit{\textbf{a}}^*$-s obtained with $\{\alpha_0, \alpha_1, ...\}$  \Comment{$\textit{\textbf{a}}^{*}_s \in \mathbb{R}^{H \times T^*}$}

\State $\overbar{\textit{\textbf{a}}}^{*}_s \gets$ $\mathbb{E}_{h} \left[ \textit{\textbf{a}}^{*}_s \right]$  \Comment{$\overbar{\textit{\textbf{a}}}^{*}_s \in \mathbb{R}^{T^*}$}

\State $\overbar{\overbar{\textit{\textbf{a}}}}^{*}_s \gets$ moving-averaged $\overbar{\textit{\textbf{a}}}^{*}_s$ with window size of $T$  \Comment{$\overbar{\overbar{\textit{\textbf{a}}}}^{*}_s \in \mathbb{R}^{T^*}$}

\State $\textit{\textbf{a}}_{\text{final}} \gets ( \overbar{\textit{\textbf{a}}}^{*}_s + \overbar{\overbar{\textit{\textbf{a}}}}^{*}_s )/2$  \Comment{final anomaly scores}
\end{algorithmic}
\end{algorithm}

We emphasize the importance of flexible $\alpha$ allowed by the masked generative modeling. Many of the existing anomaly detection methods require a fixed window size which in turn limits the performance as the smaller window size allows to capture short-ranged anomalies such as a peak but misses long-ranged anomalies such as a trend shift and vice versa. Our method, on the other hand, offers a flexible window size via $\alpha$, enabling the anomaly detection in various aspects with respect to a window size without any further training.

\subsubsection{Mapping the Anomaly Scores in the Discrete Latent Space to the Data Space}
The predicted anomaly scores in the discrete latent space $\tilde{\textit{\textbf{a}}}$ can be simply mapped to the data space using a simple nearest-neighbor interpolation technique. The resulting anomaly score is denoted by $\tilde{\textbf{\textit{a}}}_m \in \mathrm{R}^{H \times T}$.  
This interpolation involves expanding the dimension from $(H \times W)$ to $(H \times T)$, where $T$ represents the length of $\textit{\textbf{x}}$.

\subsection{Explainable Sampling}
\textit{Explainable sampling} refers to a process of masking anomalous segments in $\textit{\textbf{s}}$ and subsequently performing resampling with $p_\theta(\textit{\textbf{s}} | \textit{\textbf{s}}_M)$. 
To be more precise, we first compute the threshold as $n$-th quantile of $\textit{\textbf{a}}_{\text{final}}$ computed with a training dataset. $n$ should be determined depending on the amount of present anomalies in your dataset, and the more anomalies present in the training dataset, the lower $n$ should be. Typically, 0.9, 0.99, or 0.999 should be reasonable. 
Then, in the test dataset, we detect timesteps as anomalous if the anomaly scores $\textit{\textbf{a}}_{\text{final}}$ of the test dataset are above the threshold. We mask $\textit{\textbf{s}}$ for those anomalous timesteps across the frequency dimension, and perform $p_\theta(\textit{\textbf{s}} | \textit{\textbf{s}}_M)$ to predict the likely tokens for the masked regions. The resampling process is equivalent to the iterative decoding from TimeVQVAE. 
Fig.~\ref{fig:examples_LF_HF_anomaly_detection} presents an example of explainable sampling.
It is important to emphasize that explainable sampling provides valuable insights and interpretations for users, enhancing confidence in the model's anomaly detection. 
Further details are described in Appendix~\ref{appendix:implementation_details_timevqvae-ad}.

\section{Experiments}

\subsection{Evaluation Metric}
We follow the evaluation metric suggested by \cite{keogh2021multi}. 
In the evaluation phase, the expected output of a method is a single predicted anomaly location. If this predicted location falls within a range of ±100 data points from the true location, it is considered accurate with a accuracy score of 1.0. Conversely, if the predicted location deviates outside this range, it is considered incorrect, resulting in an accuracy score of 0.0. To provide a comprehensive assessment, the accuracies across all 250 datasets are averaged, yielding a single accuracy score.

\subsection{Experimental Setup}
All datasets from the UCR-TSA archive are used in the experiments. 
A window size is set to 2 $\times$ a period length $P$, following \cite{nakamura2023merlin++}, and each input window is z-normalized. 
For our encoder and decoder, those from TimeVQVAE are adopted and modified according to our architectural proposals, and
the vector quantizer and prior model from TimeVQVAE are adopted for TimeVQVAE-AD with a minor parameter change.
Further details on the parameter choices and implementations are available in \ref{sect:implementation_details}.

\subsection{Results}
We have reviewed the existing literature and gathered accuracy scores of various TSAD methods evaluated on the UCR-TSA archive. Furthermore, we conducted our own evaluations and obtained results for several additional competing methods. The compiled accuracy scores are presented in Table~\ref{tab:acc_table}. 
In addition to the top-1 accuracy, we report the top-k accuracies is important given that many real-world time series signals can involve multiple plausible anomalies, therefore the top-k accuracy can provide a more comprehensive evaluation of the model's performance. 
The top-3 and top-5 accuracies are measured using a simple local maxima-finding algorithm based on simple comparison of neighboring values such as the function named \texttt{find\_peaks} from the SciPy library \cite{2020SciPy-NMeth}. 
Note that the reported top-1 accuracy of COCA in its original paper is documented as 0.661. However, when we conducted experiments using the authors' GitHub codes without any modifications, our results yielded a significantly lower accuracy of 0.236. 
To ensure the validity of our finding, we compared our result with the visualization presented by the authors. Notably, the authors presented a visualization on a single specific dataset from the UCR-TSA archive. Our outcome closely aligned with their visualization.

\begin{figure}[!ht]
\centering
\subfloat{\includegraphics[width=0.47\textwidth]{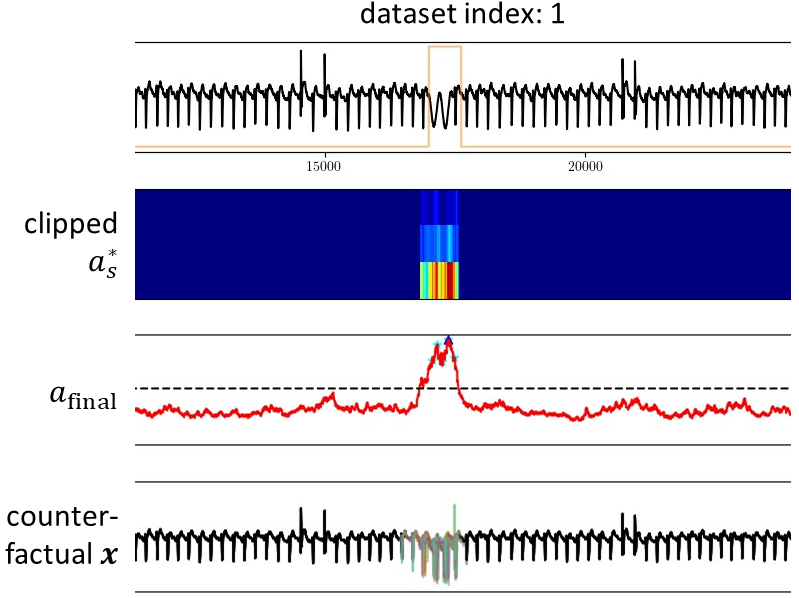}} \hspace{0.02\textwidth}
\subfloat{\includegraphics[width=0.47\textwidth]{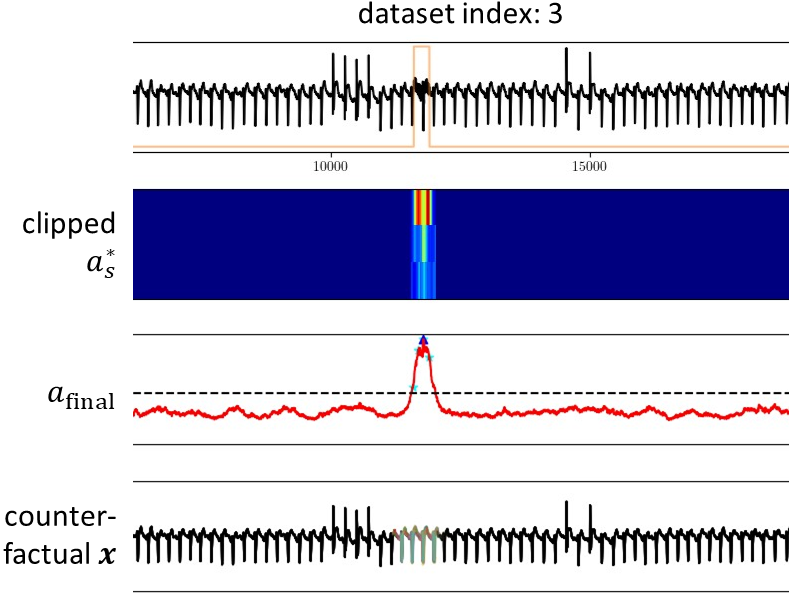}} \\ [4ex]
\subfloat{\includegraphics[width=0.47\textwidth]{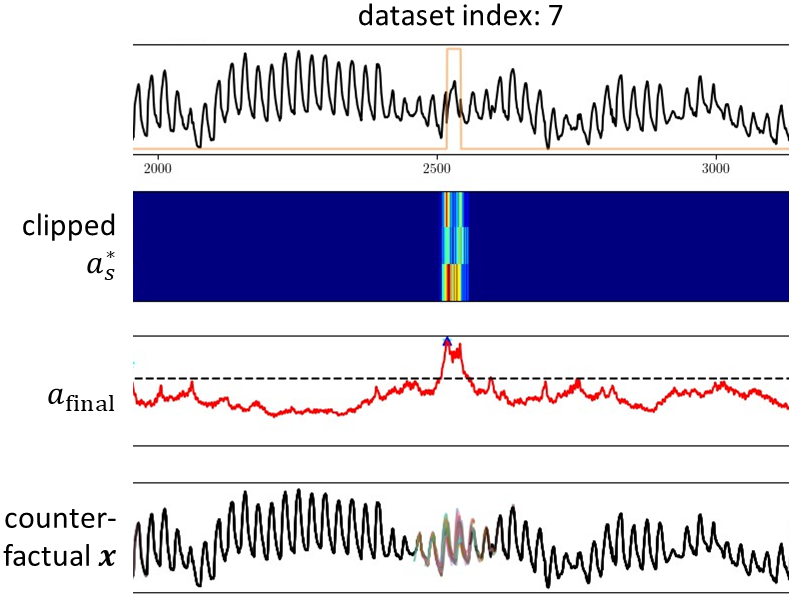}} \hspace{0.02\textwidth}
\subfloat{\includegraphics[width=0.47\textwidth]{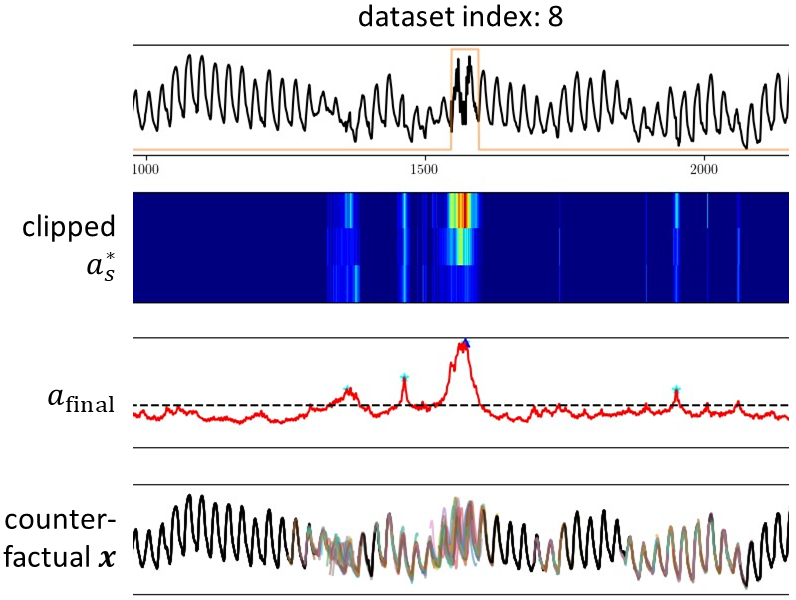}} \\ [4ex]
\subfloat{\includegraphics[width=0.47\textwidth]{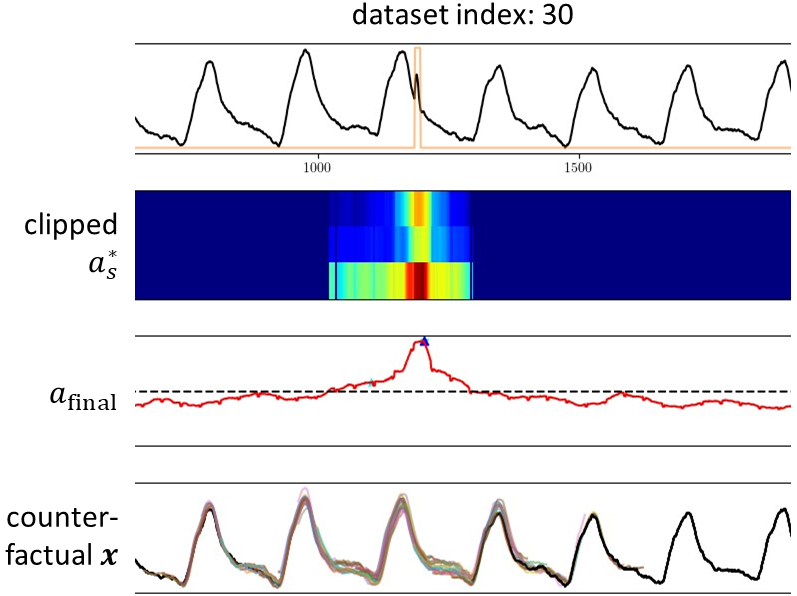}} \hspace{0.02\textwidth}
\subfloat{\includegraphics[width=0.47\textwidth]{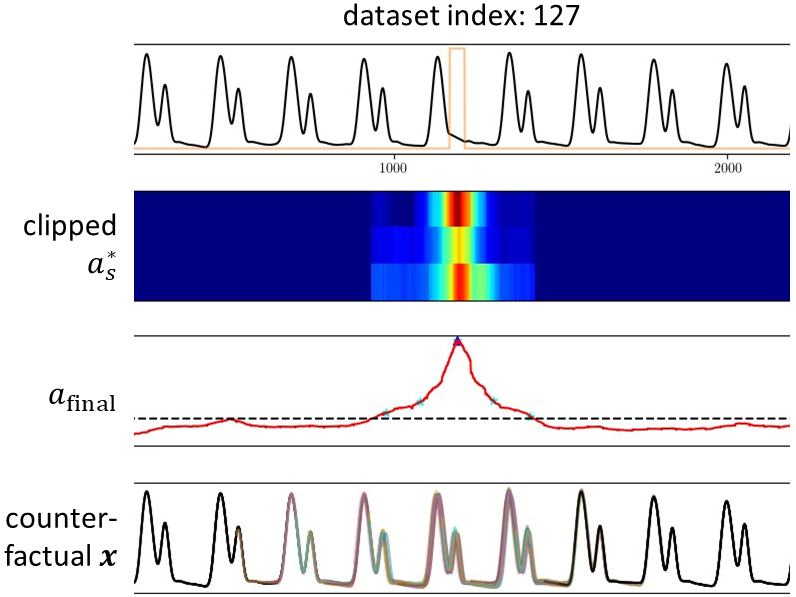}}
\caption{Examples of the visual results showcasing predicted anomaly scores by TimeVQVAE-AD on different datasets from the UCR-TSA archive. The first row shows a test time series (black) with the corresponding labels (orange), the second row presents the anomaly scores with the height representing the frequency dimension $\textit{\textbf{a}}_s^*$ clipped by the threshold, the third row presents the final anomaly scores $\textbf{\textit{a}}_{\text{final}}$, and the last row presents the likely normal states achieved through explainable sampling. 
It should be noted that the scaling of the likely normal states depends on the scaling factors of the corresponding original time series segment. For instance, if an original time series segment has a low mean value, the likely normal states also have a low mean value. The same principle applies to the scaling of the variance.}
\label{fig:result_examples_timevqvaead}
\end{figure}

Table~\ref{tab:acc_table} shows the outstanding superiority of TSAD accuracy achieved by TimeVQVAE-AD. To ensure the credibility of our results, we have made the visualizations and CSV files of the predicted anomaly scores on the UCR-TSA archive openly accessible in our GitHub repository. 
Fig.~\ref{fig:result_examples_timevqvaead} displays visualizations of predicted anomaly scores by TimeVQVAE-AD on the various datasets. 


\begin{table}[ht]
\centering
\caption{Evaluation of the various TSAD methods on the UCR-AD archive. \textit{Acc.} denotes the accuracy metric. The evaluation of the model's performance is more comprehensive when considering the top-k accuracies.
}
\label{tab:acc_table} 
\begin{tabular}{lllllll}
\toprule
 & Method Mechanism     & Method & \makecell[l]{Published \\ Year}    & \makecell[l]{Top-1\\Acc.}    & \makecell[l]{Top-3\\Acc.}    & \makecell[l]{Top-5\\Acc.} \\
 \midrule
non-DL    & one-class classification             & OC-SVM \cite{scholkopf1999support} & 1999                           & 0.088 \cite{mou2023deep} & &              \\
non-DL    & isolation forest                     & IF \cite{liu2008isolation} & 2008                            & 0.376 \cite{mou2023deep} & &              \\
non-DL    & isolation forest                     & RCF \cite{guha2016robust} & 2016                          & 0.387 \cite{mou2023deep}   & &            \\
non-DL    & matrix profile                    & Matrix Profile SCRIMP \cite{yeh2016matrix} & 2016                                           & 0.416 \cite{nakamura2023merlin++}  & &             \\
non-DL    & density estimation                   & MDI \cite{barz2018detecting} & 2018                          & 0.47 \cite{rewicki2022worth}  & &              \\
non-DL    & matrix profile                   & Matrix Profile STUMPY \cite{law2019stumpy} & 2019             & 0.512  & 0.684  & 0.744              \\
non-DL    & discord discovery                    & MERLIN \cite{nakamura2020merlin} & 2020                                           & 0.424 \cite{nakamura2023merlin++}  & &             \\
non-DL    & discord discovery                    & MERLIN++ \cite{nakamura2023merlin++} & 2023                                         & 0.424 \cite{nakamura2023merlin++}  & &             \\
\hdashline
DL        & reconstruction                       & AE & ~                                               & 0.236 \cite{audibert2021univariate}  & &             \\
DL        & reconstruction                       & Convolutional AE & ~                                 &  0.352  & 0.412 & 0.448            \\
DL        & reconstruction                       & LSTM-ED \cite{malhotra2016lstm} & 2016                                       & 0.51 \cite{mou2023deep} & &               \\
DL        & variational reconstruction           & LSTM-VAE \cite{park2018multimodal} & 2018                                         & 0.198 \cite{audibert2021univariate}  & &             \\
DL        & forecasting                          & Telemanom \cite{hundman2018detecting} & 2018                                        & 0.468 \cite{nakamura2023merlin++}  & &            \\
DL        & one-class classification             & Deep SVDD \cite{ruff2018deep} & 2018      & 0.076 \cite{mou2023deep}  & &             \\
DL        & density estimation                   & DAGMM \cite{zong2018deep} & 2018             & 0.061 \cite{mou2023deep}   & &            \\
DL        & spectral saliency map                & SR-CNN \cite{ren2019time} & 2019                   & 0.30 \cite{mou2023deep}  & &              \\
DL        & \makecell[l]{reconstruction,\\adversarial training} & USAD \cite{audibert2020usad} & 2020       & 0.276 \cite{audibert2021univariate} & &              \\
DL        & contrastive learning                 & CPC-AD \cite{de2021contrastive} & 2021           & 0.064 \cite{mou2023deep} & &              \\
DL & \makecell[l]{contrastive learning,\\one-class classification} & TS-TCC-AD \cite{eldele2021time, sohn2020learning} & 2021    & 0.006 \cite{mou2023deep} & & \\
DL        & reconstruction                       & TranAD \cite{tuli2022tranad} & 2022                           & 0.19 \cite{rewicki2022worth} & &              \\
DL        & density estimation                   & GANF \cite{dai2022graph} & 2022             & 0.24 \cite{rewicki2022worth} & &               \\
DL        & non-contrastive learning             & COCA \cite{mou2023deep} & 2023             & 0.236 & 0.328 & 0.408              \\
DL        & density estimation                   & TimeVQVAE-AD (ours) & 2023     & \textbf{0.708}  & \textbf{0.776} & \textbf{0.824}     \\              
\bottomrule
\end{tabular}  
\end{table}

\section{Discussion}
\label{sect:discussion}
\paragraph{How is TimeVQVAE-AD Significantly Better than the Others?}

We employ the strongest prior model for time series generation from TimeVQVAE. This prior model can generate synthetic time series with high fidelity, indicating its effectiveness in approximating a target prior. 
By utilizing this strong prior model in measuring anomaly scores, we are able to achieve more accurate anomaly assessments.
Importantly, our anomaly scores do not rely on the error between a target time series and predicted time series, unlike the majority of existing TSAD methods that do rely on such error. Typically, those methods are limited to detecting anomalies with abnormally-high amplitudes. In contrast, our anomaly scores are measured using the prior model, which captures the semantic relationships between different time series segments represented by distinct tokens. This approach enables us to effectively measure anomaly scores and detect anomalies beyond just amplitude abnormalities.
Moreover, TimeVQVAE-AD enhances its detection accuracy by measuring anomaly scores in various aspects, especially across different frequency bands and various latent window sizes. 
Regarding explainability, our proposed encoder retains the semantics of both temporal and frequency dimensions, allowing the resulting anomaly scores to carry both dimensions. Notably, the inclusion of the frequency dimension is novel in TSAD and offers valuable diagnostic insights. Additionally, the generative nature of the prior model enables resampling anomalous segments and obtain corresponding likely normal states. These two properties significantly enhance the confidence in the detected anomalies.

\begin{figure}[!ht]
\centering
\includegraphics[width=0.45\textwidth]{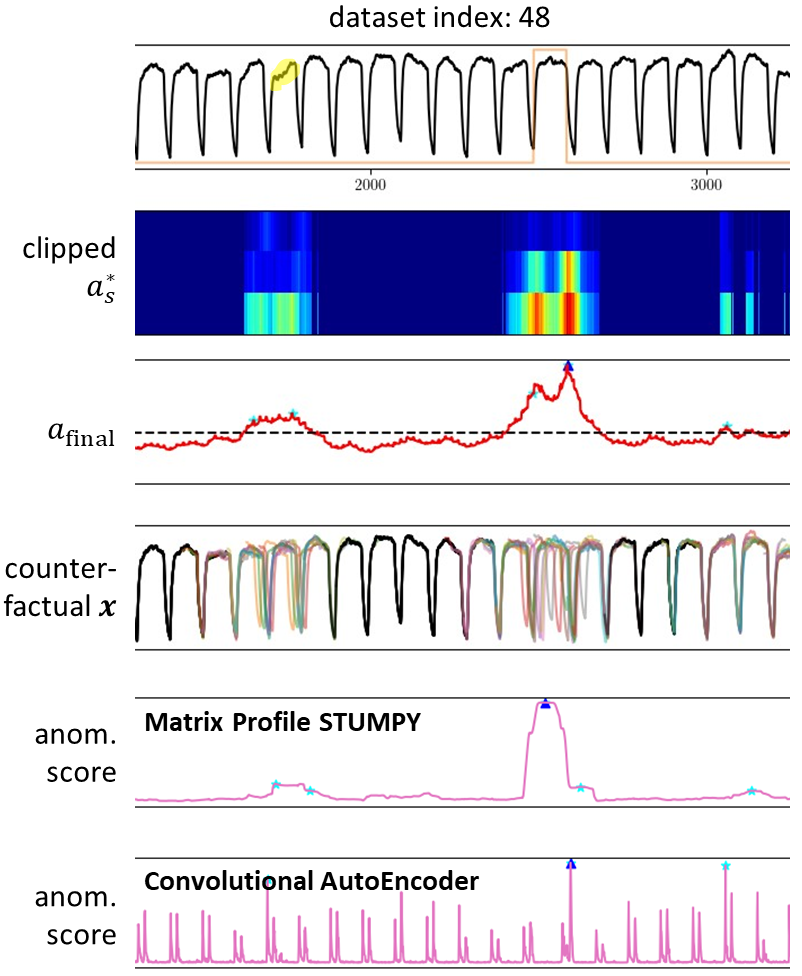}
\caption{Example of a time series dataset with a definite anomaly (labeled in orange) and subtle anomaly (colored in yellow) with corresponding predicted anomaly scores by TimeVQVAE-AD, Matrix Profile STUMPY, and Convolutional AE. 
The labeled anomaly exhibits an abnormality in its period length, which is slightly longer than what is considered normal. 
It should be noted that TimeVQVAE-AD assigns the highest scores to the labeled segment and sufficiently-high scores to the subtle anomaly. 
Moreover, TimeVQVAE-AD demonstrates a more precise capture of the labeled anomaly in comparison to STUMPY. Given that the anomaly is characterized by an abnormal period, it is essential that both ends of the anomaly are accurately identified as the most anomalous timesteps, which is effectively accomplished by TimeVQVAE-AD.
}
\label{fig:discussion-magnitude_of_anomalism}
\end{figure}

\paragraph{Anomaly Scores Should Measure Magnitude of Anomalism}
Anomalism refers to the quality of being anomalous. Anomalies can exhibit varying degrees of magnitude, ranging from slightly anomalous to moderately anomalous and completely anomalous. A robust TSAD method should be capable of effectively capturing this spectrum. 
In Fig.~\ref{fig:discussion-magnitude_of_anomalism}, we present an example showcasing a time series with a definite anomaly (labeled in orange) and subtle anomaly (yellow) with corresponding predicted anomaly scores by TimeVQVAE-AD, Matrix Profile STUMPY, and Convolutional AE.
TimeVQVAE-AD and STUMPY successfully assign the highest scores to the labeled segment and sufficiently-high scores to the subtle anomaly, while TimeVQVAE-AD achieves a more precise capture of the anomaly. 
However, Convolutional~AE fails. This is because Convolutional~AE is regularized due to its bottleneck during training and can effectively reconstruct a time series from a test dataset that closely resembles those in the training dataset. Consequently, the reconstruction error remains low for the subtle anomaly due to the effective  reconstruction.
Regarding TimeVQVAE-AD, it is important to acknowledge that the prior model learns the prior distribution, which in turn allows for an effective capture of magnitude of anomalism.

\begin{figure}[!ht]
\centering
\subfloat{\includegraphics[width=0.47\textwidth]{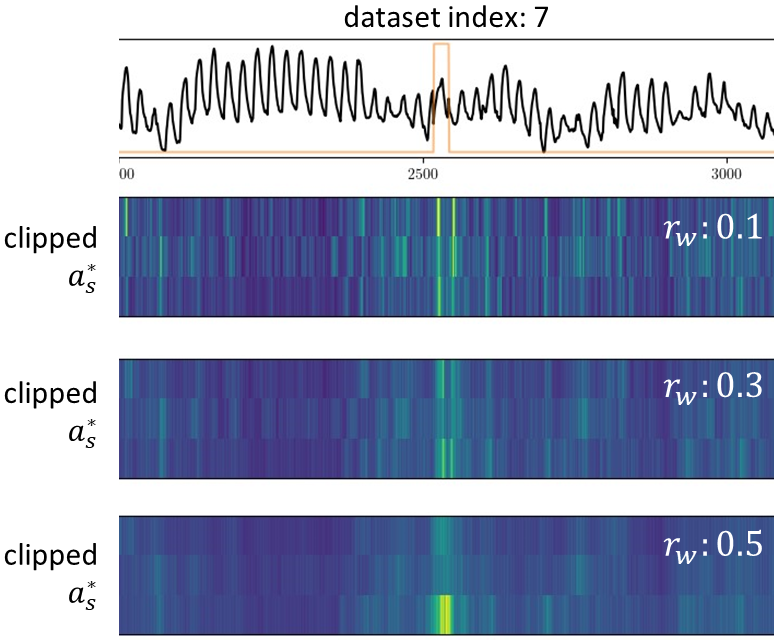}}  \hspace{0.03\textwidth} 
\subfloat{\includegraphics[width=0.47\textwidth]{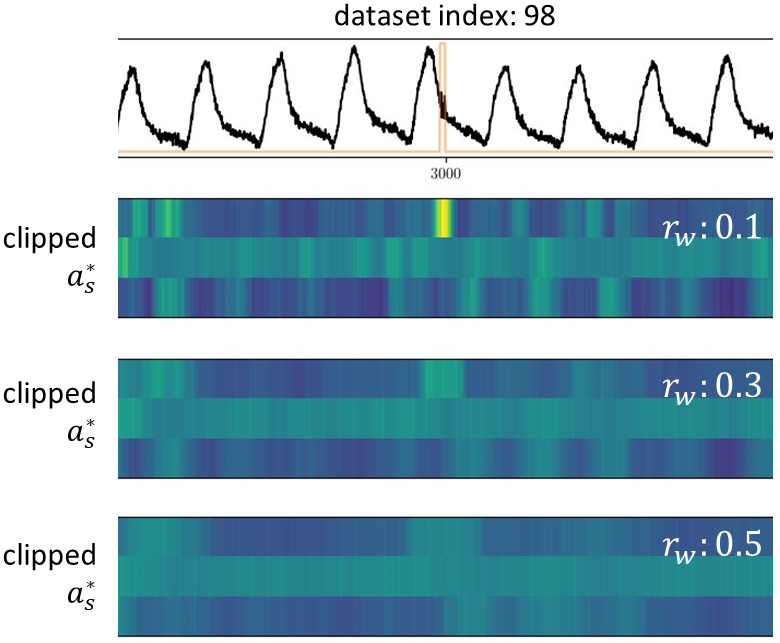}}
\caption{Examples of the predicted anomaly scores with different latent window sizes determined by the latent window size rate $r_w$. The first example involves a long-range anomaly and the second one involves a short-range anomaly. 
It is noticeable that the long-range anomaly is better captured with a wider latent window size rate, while the short-range anomaly is better detected with a narrower latent window size rate.}
\label{fig:discussion-window_size}
\end{figure}

\paragraph{Flexible Window Size Enabled by $\alpha$}
Unlike the existing deep TSAD methods, TimeVQVAE-AD allows a flexible window size, enabled by $\alpha$. It defines the temporal window range in the discrete latent space as $[w-\alpha, w+\alpha]$ which is analogically equivalent to a kernel size of a convolutional layer. 
The captured anomaly aspects vary depending on the latent window size. A narrow window size is effective in detecting short-range anomalies, while a wide window size is effective in capturing long-range anomalies.
Fig.~\ref{fig:discussion-window_size} presents examples of the predicted anomaly scores $\textit{\textbf{a}}^*$ with different latent window sizes. In practice, the latent window size is set as $r_w \times W$ where $r_w$ denotes a latent window size rate and $r_w \in (0, 1)$, and we use $r_w$ of $\{0.1, 0.3, 0.5\}$ in our experiments to cover from a narrow to wide window.
We must emphasize that the incorporation of the effects of different window sizes enhances our detection accuracy.

\begin{figure}[!ht]
\centering
\subfloat{\includegraphics[width=0.44\textwidth]{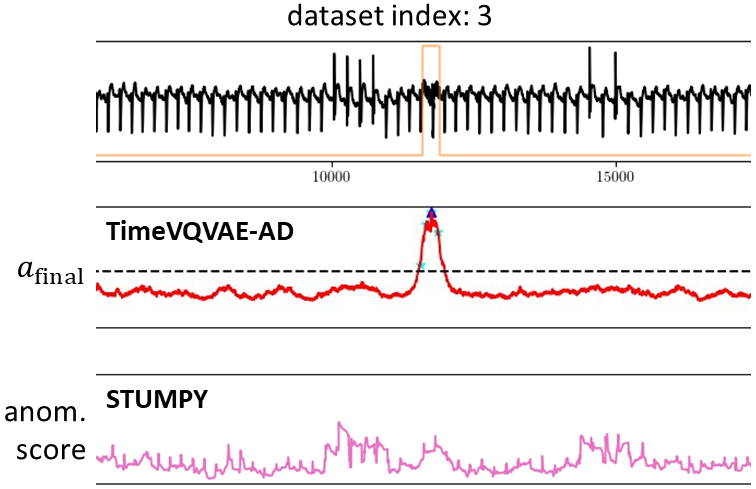}}  \hspace{0.03\textwidth}
\subfloat{\includegraphics[width=0.44\textwidth]{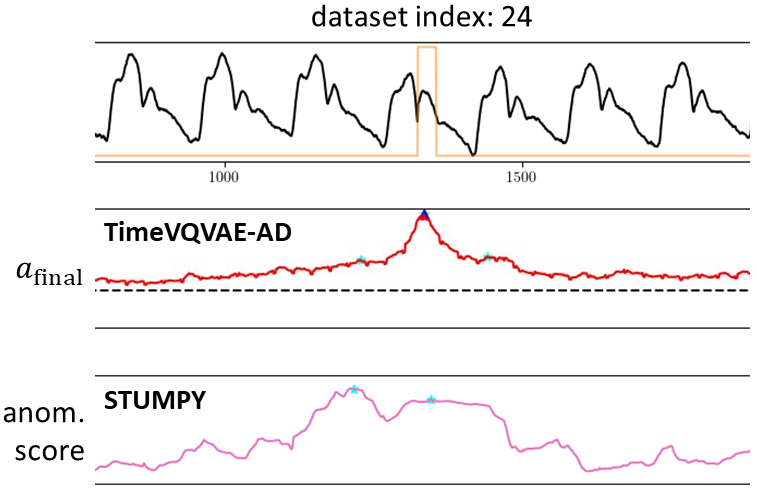}}
\caption{Example of time series datasets where Matrix Profile STUMPY fails. It highlights the inherent limitations associated with matrix profile and discord discovery methods for TSAD.}
\label{fig:discussion-where_STUMPY_fails}
\end{figure}

\paragraph{Where Do Matrix Profile and Discord Discovery Methods Fail?}
Table~\ref{tab:acc_table} demonstrates that, in general, the matrix profile and discord discovery methods outperform the existing deep learning methods in terms of detection accuracy. Yet, the best accuracy achieved by those methods is 0.512, achieved by Matrix Profile STUMPY. This raises the question of where those methods fail.
Both methods primarily rely on measuring anomaly scores by evaluating the distances between different subsequences of time series. Consequently, they face a similar challenge to the reconstruction and forecasting-based TSAD methods -- that is, they can typically capture anomalies with abnormally-high amplitudes only since they measure anomaly scores based on the error between a target time series and its predicted counterpart.
Fig.~\ref{fig:discussion-where_STUMPY_fails} presents an example of time series datasets where Matrix Profile STUMPY fails due to its limitation, whereas TimeVQVAE-AD succeeds.

\begin{figure}[!ht]
\centering
\subfloat{\includegraphics[width=\textwidth]{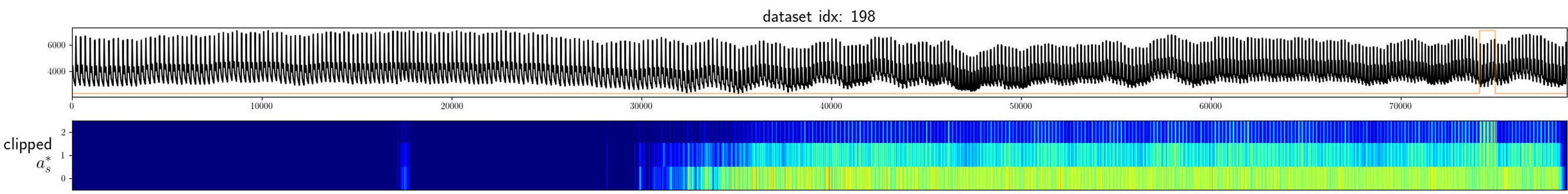}}\\[1.5ex]
\subfloat{\includegraphics[width=\textwidth]{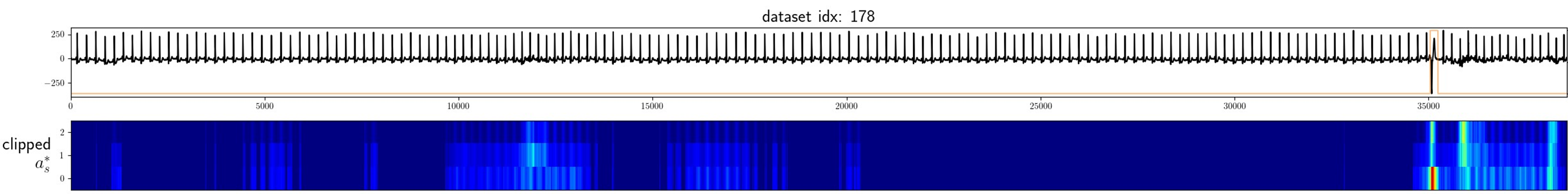}}\\[1.5ex]
\subfloat{\includegraphics[width=\textwidth]{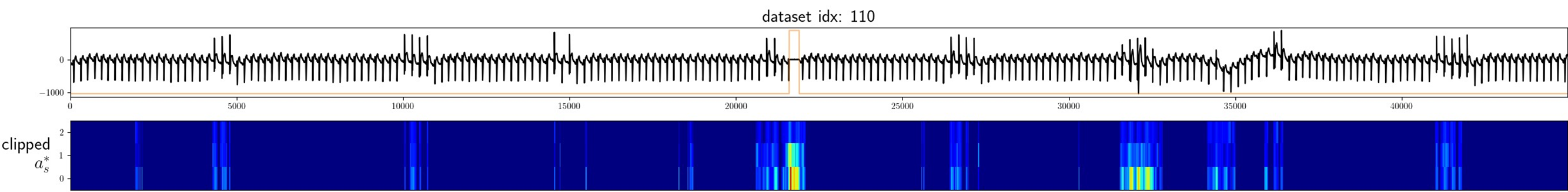}}
\caption{Examples of the datasets with hidden anomalies (\textit{i.e.,} unlabeled anomalies), detected by TimeVQVAE-AD. 
In the first dataset, we notice an increase in density towards the right side. 
In the second dataset, upon closer examination, we can observe the anomalous patterns on the right side. 
The third dataset reveals hidden anomalies in the form of unusual amplitudes occurring around 32000.} 
\label{fig:discussion-hidden_anomalies}
\end{figure}

\paragraph{Presence of Hidden Anomalies}
We have identified that there exist hidden anomalies, in other words, unlabeled anomalies. Perhaps that is natural to observe, referring to the argument from the author of the UCR-TSA archive, "In fact, perfect ground truth labels are impossible for anomaly detection" \cite{no_perfect_label}. 
Fig.~\ref{fig:discussion-hidden_anomalies} presents examples of the datasets with hidden anomalies, detected by TimeVQVAE-AD.
The predicted anomaly scores on such hidden anomalies can obscure the scores on labeled anomalies, leading to the accuracy of 0. However, it is worth considering that the model may have successfully detected the labeled anomalies with the second or third highest scores. Therefore, the evaluation of TSAD should also account for this capability. To accommodate for such scenarios, we suggest the adoption of top-k accuracies.

\begin{figure}[!ht]
\centering
\subfloat{\includegraphics[width=0.42\textwidth]{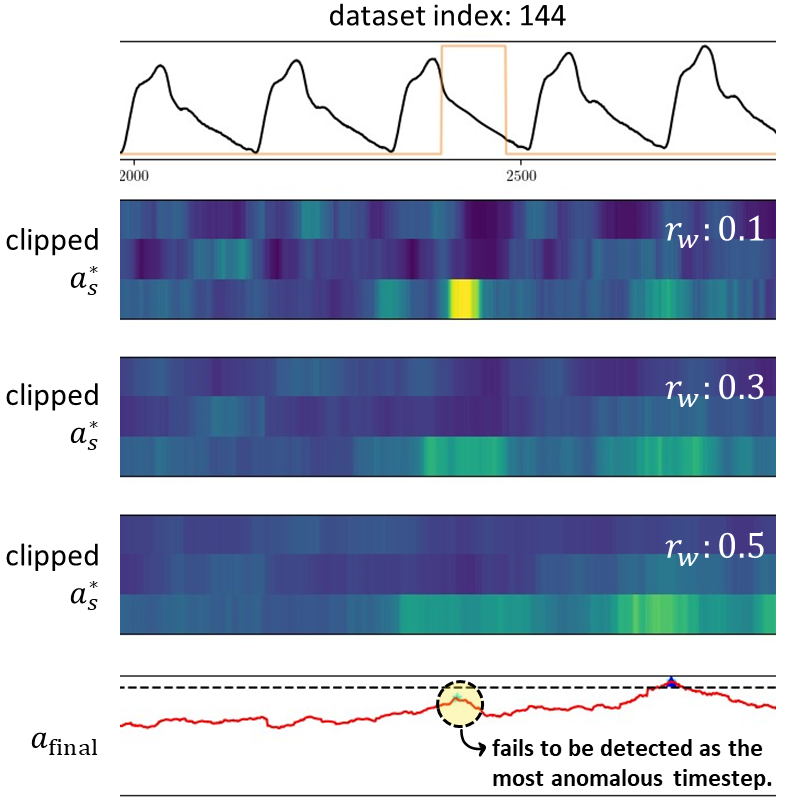}}  \hspace{0.03\textwidth} 
\subfloat{\includegraphics[width=0.42\textwidth]{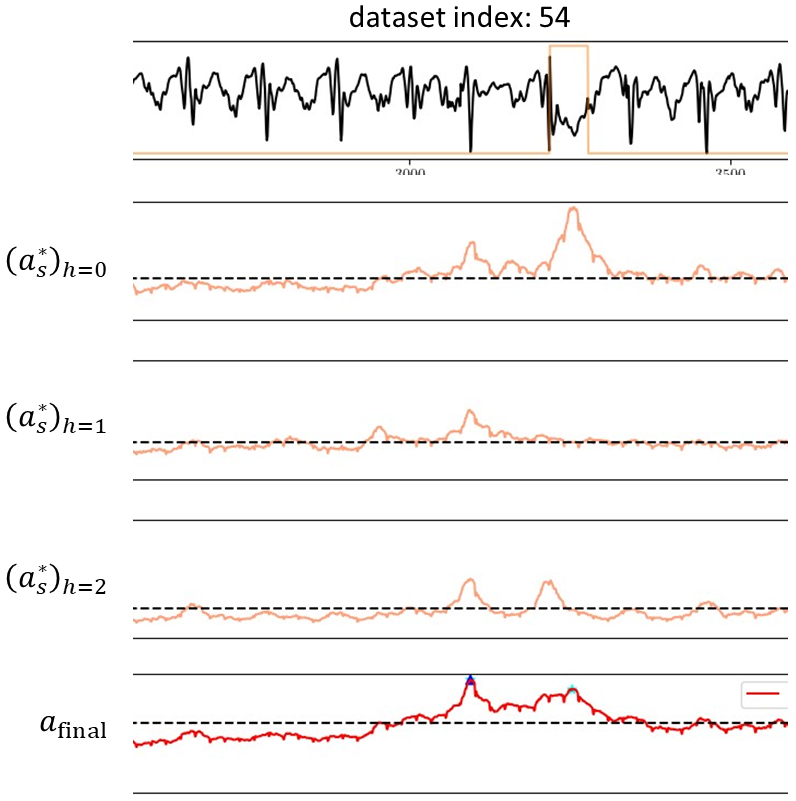}}
\caption{Examples of the dilemma regarding the aggregation of predicted anomaly scores. 
$h$ denotes a frequency band, and $h=0$, $h=1$, and $h=2$ denote the lowest, middle, and highest frequency bands, respectively.
The left figure illustrates a scenario in which the anomaly is effectively captured using a small latent window size, but fails to be detected with larger latent window sizes, ultimately resulting in a detection failure.
The right figure illustrates a scenario in which the anomaly score displays a prominent peak in one frequency band while remaining low in other frequency bands, leading to a detection failure despite the anomaly's evident presence.
}
\label{fig:discussion-aggregation}
\end{figure}

\paragraph{Dilemma of Aggregation of Anomaly Scores}
The predicted anomaly scores of TimeVQVAE-AD capture two novel aspects: 1) different frequency bands and 2) different temporal segment sizes determined by $\alpha$. 
Therefore, we have $n$ number of anomaly scores $\textit{\textbf{a}}_s^*$ before the aggregation process, where $n$ is $\text{a number of $\alpha$-s}$.
These anomaly scores, however, must be aggregated to yield a single value for each timestep, as required for the computation of the evaluation metrics. Unfortunately, this aggregation process can sometimes compromise the metrics scores.
For example, when an anomaly has a short duration, it is effectively captured by the anomaly score associated with a small $\alpha$, but it is less likely to be detected by the score linked to a larger $\alpha$. As a result, after the aggregation, the anomaly score with a small $\alpha$ may become too attenuated to emerge as the highest anomaly score.  
Another failure scenario arises when an anomaly score exhibits a distinctive peak in one frequency band but remains low in the other frequency bands. In such cases, the anomaly might fail to be identified as the top anomaly after the aggregation process for the same reason as above. We note, however, that a practitioner may choose to look at anomaly scores in certain frequency bands rather than at aggregated level depending on the application at hand. While the aggregated score may fail to detect an anomaly according to our proposed scheme, a practitioner has the opportunity to look at different frequency bands for further analysis. 
Examples of the first and second scenarios are shown in Fig.~\ref{fig:discussion-aggregation}.

\begin{figure}[!ht]
\centering
\includegraphics[width=0.6\textwidth]{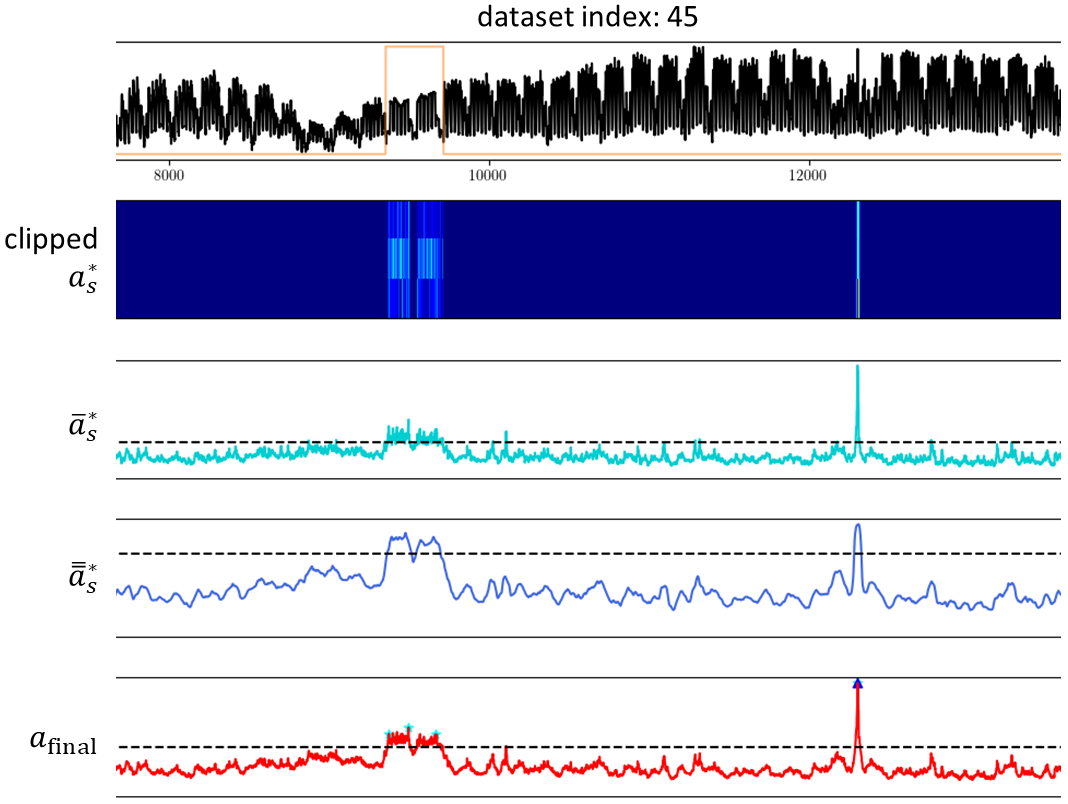}
\caption{Example of the dilemma regarding long-range medium-amplitude anomaly scores. The anomaly segment at around 9600 has long-range medium-amplitude anomaly scores, while there are two other short-range high-amplitude anomaly segments. 
The dilemma is that the significance of long-range medium-amplitude anomaly scores is not properly accounted for. To mitigate the dilemma, moving-average is applied to $\bar{\textit{\textbf{a}}}_s^*$, producing $\bar{\bar{\textit{\textbf{a}}}}_s^*$ and it is noticeable that $\bar{\bar{\textit{\textbf{a}}}}_s^*$ better captures the significance of long-range medium-amplitude anomaly scores. 
}
\label{fig:discussion-dilemma_long_range_medium_amplitude}
\end{figure}

\paragraph{Dilemma of Long-range Medium-amplitude Anomaly Scores}
In physics, there is a physical quantity called \textit{impulse} $J$, defined as $J = F \times \Delta{t}$ where $F$ and $\Delta{t}$ denote force and duration. That indicates that $J$ is large even if $F$ is moderate as long as $\Delta{t}$ is sufficiently large. 
The concept of $J$ can be applied to anomaly scores. For instance, if an anomalous segment has a long range with medium-amplitude scores, it should still be regarded as highly anomalous due to its extensive duration.
However, in the evaluation protocol, a method is typically required to return a single most-likely anomalous timestep, often obtained through an argmax operation. This approach does not adequately account for the significance of long-range medium-amplitude anomaly scores.
To mitigate the limitation, we introduce the concept of $\Delta{t}$ in calculating $\textit{\textbf{a}}_{\text{final}}$. 
As described in Algorithm~\ref{alg:anomaly_score_prediction}, we propose applying moving-average with a window size of $T$ to $\bar{\textit{\textbf{a}}}_s^*$, resulting in $\bar{\bar{\textit{\textbf{a}}}}_s^*$. Conceptually, $\bar{\bar{\textit{\textbf{a}}}}_s^*$ corresponds to $F \times \Delta{t}$, while $\bar{\textit{\textbf{a}}}_s^*$ corresponds to $F$. However, $\bar{\bar{\textit{\textbf{a}}}}_s^*$ loses locality, so we combine $\bar{\textit{\textbf{a}}}_s^*$ and $\bar{\bar{\textit{\textbf{a}}}}_s^*$ to obtain $\textit{\textbf{a}}_{\text{final}}$.
Fig.~\ref{fig:discussion-dilemma_long_range_medium_amplitude} illustrates an example that highlights the dilemma. 

\section{Conclusion}
In this paper, we touched upon significant flaws in the currently-popular benchmark datasets and evaluation protocol and limitations of the existing TSAD methods in terms of detection accuracy and explainability. 
Then we proposed TimeVQVAE-AD, a novel approach to TSAD that leverages a strong prior model from the SOTA time series generation method, TimeVQVAE, to learn the distributions of likely, less likely, and unlikely normal states of time series with respect to different frequency bands.
This distribution modeling enables the detection of a broad spectrum of anomalies across different frequency bands and temporal scales, which in turn allows for the detection of subtle and complex anomalies that may be overlooked by traditional TSAD methods. Moreover, our method's focus on explainability through the generation of counterfactual examples provides valuable insights into the detected anomalies.
Our experiments were conducted on the UCR Time Series Anomaly archive for a fair evaluation. The experimental results showed that TimeVQVAE-AD achieves ground-breaking TSAD accuracy and an exceptional level of explainability through counterfactual samples. 

Our work opens several paths for future research. Firstly, improving TimeVQVAE-AD involves overcoming the challenges in anomaly score aggregation and detecting long-range medium-amplitude anomalies. Secondly, applying our method to multivariate time series could greatly expand its use and impact. Additionally, exploring its potential across different sectors such as healthcare and finance could not only validate its effectiveness in real-world scenarios but also lead to new insights and improvements in the methodology.
Overall, we believe TimeVQVAE-AD marks a major progress in the field of TSAD.

\section*{Acknowledgments}
We would like to thank the Norwegian Research Council for funding the Machine Learning for Irregular Time Series (ML4ITS) project (312062). This funding directly supported this research.
We also would like to thank the people who have contributed to the UCR Time Series Anomaly archive \cite{wu2021current} for their valuable contribution to the AD community.

\section*{Ethical Statement}
No conflicts of interest were present during the research process.

\bibliographystyle{unsrt}  
\bibliography{references}  

\newpage
\appendix

\section{Implementation Details}
\label{sect:implementation_details}

\subsection{Dataset}
All datasets from the UCR-TSA archive are utilized in our experiments. The window size is set to $2 \times P$ in accordance with \cite{nakamura2023merlin++}, and each input window underwent z-normalization. To ensure comprehensive coverage of time series patterns, we set the window stride to 1, avoiding the possibility of missing any specific pattern.
Regarding the parameter $P$, we manually measured its value for all 250 datasets and made the measurements available in our GitHub repository. While \cite{nakamura2023merlin++} employed an autocorrelation function to determine $P$, we observed that the periods of several datasets presented challenges when using autocorrelation. Consequently, in an effort to establish a standardized evaluation protocol, we conducted meticulous manual measurements of $P$ for all datasets.

\subsection{TimeVQVAE-AD}  
\label{appendix:implementation_details_timevqvae-ad}
\paragraph{STFT, ISTFT}
STFT and ISTFT are implemented with \texttt{torch.stft} and \texttt{torch.istft}, respectively. Their main parameter -- \texttt{n\_fft}, size of Fourier transform -- determines the frequency dimension size. The frequency dimension of $\mathrm{STFT}(\textit{\textbf{x}})$ is determined by $\lfloor {\texttt{n\_fft} / 2} \rfloor + 1$. 
Consequently, a higher value of \texttt{n\_fft} leads to anomaly scores with a larger frequency dimension, enabling the detection of anomalies with finer frequency resolution.
Through our experiments, we found that an \texttt{n\_fft} value of 4 was sufficient to capture various types of anomalies present in the UCR-TSA archive.

\paragraph{Encoder}
The encoder used in TimeVQVAE-AD shares the same structure as that of TimeVQVAE, with the exception of kernel size modifications and the utilization of a smaller compression rate, as detailed in Sect.~\ref{sect:architecture}. The compression rate can be alternatively specified by \textit{downsampled width} \cite{lee2023vector}, which refers to the width of $\textit{\textbf{z}}_q$. A smaller downsampled width corresponds to a higher compression rate, and increasing the number of downsampling blocks in the encoder achieves a higher compression rate. In TimeVQVAE-AD, we employ a downsampled width of 32. 
The selection of the downsampled width is important. When opting for a smaller downsampled width, it results in lower resolution in the prediction of AD scores since an element of $z_q$ encompasses a broader temporal segment. Conversely, opting for a larger downsampled width can potentially lead to a reduction in the accuracy of AD predictions, as the number of tokens to predict increases for the same temporal segment. Thus, we chose 32 as a good trade-off between the prediction resolution and AD prediction accuracy in our experiments.

\paragraph{Vector Quantizer}
The same vector quantizer from TimeVQVAE is used except for the use of a bigger codebook size (128) to better capture different patterns in time series data. Also, in stage~2, we train the code embeddings fresh without re-initialization with the learned codes in stage~1 because we have a large sum of training data.

\paragraph{Decoder}
The same decoder architecture from TimeVQVAE is used.

\paragraph{Prior Model} 
The prior model from TimeVQVAE is used with a slight modification in the output layer. The output of the prior model in TimeVQVAE is computed by forming a covariance matrix between the output from the transformer model and the codebook embeddings. 
In our implementation, we simply use a shallow network to project the transformer's output to logits over the codebook.

\paragraph{Anomaly Score Prediction}
In the anomaly score prediction, there exists one hyperparameter that needs to be determined, which is a set of $\alpha$ values. Each $\alpha$ determines a latent window size. As explained in Sect.~\ref{sect:discussion}, a latent window size can be alternatively defined as $r_w \times W$, where $r_w$ represents a latent window size rate and $r_w \in (0, 1)$. In our experiments, we cover a range of latent window sizes from narrow to wide by using $r_w$ values of ${0.1, 0.3, 0.5}$. To guarantee a minimum context for anomaly score prediction, we refrain from using $\alpha$ exceeding 0.7.
Another parameter is stride size of a rolling window. A rolling window refers to a window from $t$ to $t+T$, as specified in $\textit{\textbf{x}}^*_{t:t+T}$ in Algorithm~\ref{alg:anomaly_score_prediction}. Thus, the size of the rolling window is $T$. In the pseudocode, the stride is set to 1, indicating the rolling window shifts from left to right by $t$ of 1. 
For the faster inference, the stride can be larger than 1 such as $\lfloor r_\text{rolling.window.stride} \times T \rceil$ where $r_\text{rolling.window.stride}$ denotes a rolling window stride rate with its valid range between 0 and 1. 
Importantly, $r_\text{rolling.window.stride}$ involves a trade-off between the detection accuracy and inference speed-up. A large value of the rate leads to fast inference but can hinder the detection accuracy and vice versa. 
In our experiments, we set $r_\text{rolling.window.stride}$ to 0.1 unless specified otherwise.

\paragraph{Explainable Sampling} 
To perform the explainable sampling, the prior model employs an iterative decoding sampling approach \cite{chang2022maskgit, lee2023vector,lee2023masked}, which involves a hyperparameter known as the number of decoding steps denoted as $T_s$. In our implementation, we set $T_s$ to 20, allowing for finer sampling compared to the default value of 10 used in TimeVQVAE.
Additionally, we introduce another hyperparameter called \textit{maximum masking rate for explainable sampling}. During explainable sampling, we begin by masking anomalous segments in $\textit{\textbf{s}}$. However, there is a possibility of completely masking $\textit{\textbf{s}}$ if all timesteps have anomaly scores above the threshold. In such cases, the prior model ends up performing unconditional sampling due to the absence of contextual information. To ensure a minimum context for explainable sampling, we limit the masking rate to a maximum of 90\% of the temporal dimension, ensuring that a portion of $\textit{\textbf{s}}$ remains unmasked.

\paragraph{Optimizer}
The AdamW optimizer \cite{loshchilov2017decoupled} is used with specific settings: a batch size of 512 for stage~1 (256 if memory constraints arise) and a batch size of 512 for stage~2 (256 if memory constraints arise). The initial learning rate is set to 1e-3, and a cosine scheduler is employed as the learning rate scheduler. Additionally, a weight decay of 1e-5 is applied. Regarding the maximum number of epochs, 500 epochs for stage~1 and 1,000 epochs for stage~2 are used. Training ends if the duration exceeds 12 hours.
The training is performed with GeForce GTX 1080 Ti.

\subsection{Convolutional AutoEncoder}
Convolutional AE is one of the competing methods presented in Table~\ref{tab:acc_table}. It is implemented by utilizing the encoder and decoder architectures from TimeVQVAE-AD and replacing the existing layers with their corresponding one-dimensional versions.

\section{Faster Inference}
\begin{figure}[!ht]
\centering
\includegraphics[width=0.8\textwidth]{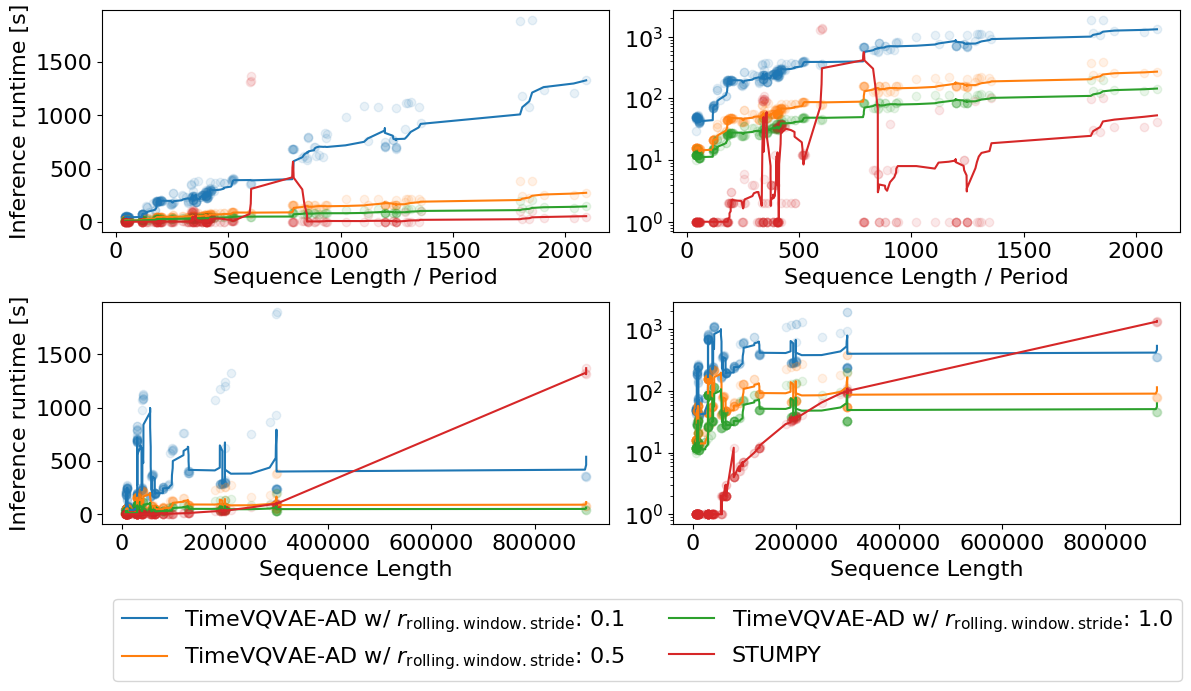}
\caption{Inference runtime vs sequence length (and sequence length/period). 
The first and second columns present a normal time scale and log time scale in the y-axis, respectively, and the first and second rows present (sequence length / period) and sequence length, respectively.
TimeVQVAE-AD with different values of $r_\text{rolling.window.stride}$ and STUMPY, the second best performing model, are compared.}
\label{fig:runtime_vs_seq/period}
\end{figure}

Inference speed is important from the operational perspective of the method. The inference runtime is determined by the time spent on running Algorithm~\ref{alg:anomaly_score_prediction}. There are two lines in the pseudocode that can allow inference speed-up: 1) $\text{\textbf{for}} \; \alpha \in \{ \alpha_0, \alpha_1, ... \} \; \text{\textbf{do}}$ and 2) $\text{\textbf{for}} \; t \in [0,1,2,...] \; \text{\textbf{do}}$. The former can be computed in parallel using multi-processing and the latter can enable the speed-up by having a stride size of $\lfloor r_\text{rolling.window.stride} \times T \rceil$. While the multi-processing on $\alpha$ does not affect the detection accuracy, the stride size involves a trade-off between detection accuracy and inference speed.
Table~\ref{tab:accuracy_and_runtime_wrt_rolling_window_stride_rate} presents the top-k accuracies and total inference runtime for the 250 datasets.
Moreover, Fig.~\ref{fig:runtime_vs_seq/period} presents the runtime for different sequence lengths from different datasets. To be more precise, sequence length and (sequence length / period $P$ = number of periods in a sequence) are presented in the x-axis. 
These experiments were performed with Intel Core i9 11900K for CPU and GeForce RTX 3060 for GPU.

Table~\ref{tab:accuracy_and_runtime_wrt_rolling_window_stride_rate} demonstrates that the trade-off between the accuracy and computational efficiency is clearly observed. While the rate of 0.1 achieves the best accuracy, the rate of 0.5 achieves accuracy close to the best one with a significantly smaller runtime, which makes it perhaps more desirable in practice. Furthermore, Fig.~\ref{fig:runtime_vs_seq/period} reveals a linear runtime increase for TimeVQVAE-AD whereas a quadratic increase for STUMPY. 
For a sequence with $N$ windows (\textit{i.e.,} $N$ subsequences), TimeVQVAE-AD runs the inference process of the prior model by $N$ times only  (\textit{i.e.,} $O(N)$). On the other hand, a Matrix Profile method needs to compute the pairwise distances between all subsequences, resulting in $O(N^2)$ pairwise distance calculations.

\begin{table}[!ht]
    \caption{Accuracy and total inference runtime with respect to $r_\text{rolling.window.stride}$ values of 0.1, 0.5, and 1.0, respectively.}
    \label{tab:accuracy_and_runtime_wrt_rolling_window_stride_rate}
    
    \centering
    \begin{tabular}{ccccc}
    \toprule
    $r_\text{rolling.window.stride}$ & Top-1 Acc. & Top-3 Acc. & Top-5 Acc. & Total inference runtime [s] \\
    \midrule
    0.1 & 0.708 & 0.776 & 0.824 & 81609  \\
    0.5 & 0.7 & 0.784  & 0.8  & 18262 \\
    1.0 & 0.632 & 0.772 & 0.788 & 10352 \\
    \bottomrule
    \end{tabular}
\end{table}

\end{document}